\documentclass[12pt,a4paper]{article}

\usepackage{amsthm}
\usepackage[british]{babel}

\usepackage[a4paper,top=2cm,bottom=2cm,left=2.5cm,right=2.5cm,marginparwidth=1.75cm]{geometry}

\newtheorem{definition}{Definition}%
\usepackage{setspace}
\usepackage{microtype}
\usepackage{graphicx}
\usepackage{float}
\usepackage{booktabs} 
\usepackage{amsmath,amsfonts}
\usepackage{bbm}
\usepackage{csquotes}
\usepackage{subcaption}
\usepackage{enumitem}
\usepackage{multirow}

\usepackage{hyperref}

\usepackage[T1]{fontenc}
\usepackage{authblk}

\usepackage{algorithm,algpseudocode}

\usepackage[authoryear]{natbib}
\begin{document}

\title{Understanding Prediction Discrepancies in Classification}

\author[1,2,*]{Xavier Renard}
\author[1,2,*]{Thibault Laugel}
\author[1,2,3]{Marcin Detyniecki}
\affil[1]{\small AXA, Paris, France}
\affil[2]{\small TRAIL, Sorbonne University, CNRS, LIP6, Paris, France}
\affil[3]{\small Polish Academy of Science, IBS PAN, Warsaw, Poland}
\affil[*]{\small Equal contribution and corresponding authors: \texttt{\{xavier.renard, thibault.laugel\}@axa.com}}

\date{}

\maketitle

\abstract{
    A multitude of classifiers can be trained on the same data to achieve similar performances during test time while having learned significantly different classification patterns.
    When selecting a classifier, the machine learning practitioner has no understanding on the differences between models, their limits, where they agree and where they don't. But this choice will result in concrete consequences for instances to be classified in the discrepancy zone, since the final decision will be based on the selected classification pattern. Besides the arbitrary nature of the result, a bad choice could have further negative consequences such as loss of opportunity or lack of fairness.
    This paper proposes to address this question by analyzing the prediction discrepancies in a pool of best-performing models trained on the same data.
    A model-agnostic algorithm, DIG, is proposed to \emph{capture and explain} discrepancies locally in tabular datasets, to enable the practitioner to make the best educated decision when selecting a model by anticipating its potential undesired consequences.}


\paragraph{\small Code access:}
{\url{https://github.com/axa-rev-research/discrepancies-in-machine-learning}}

\section{Introduction}
\label{sec:introduction}

\begin{figure}[t]
    \centering
    \includegraphics[width=0.75\linewidth]{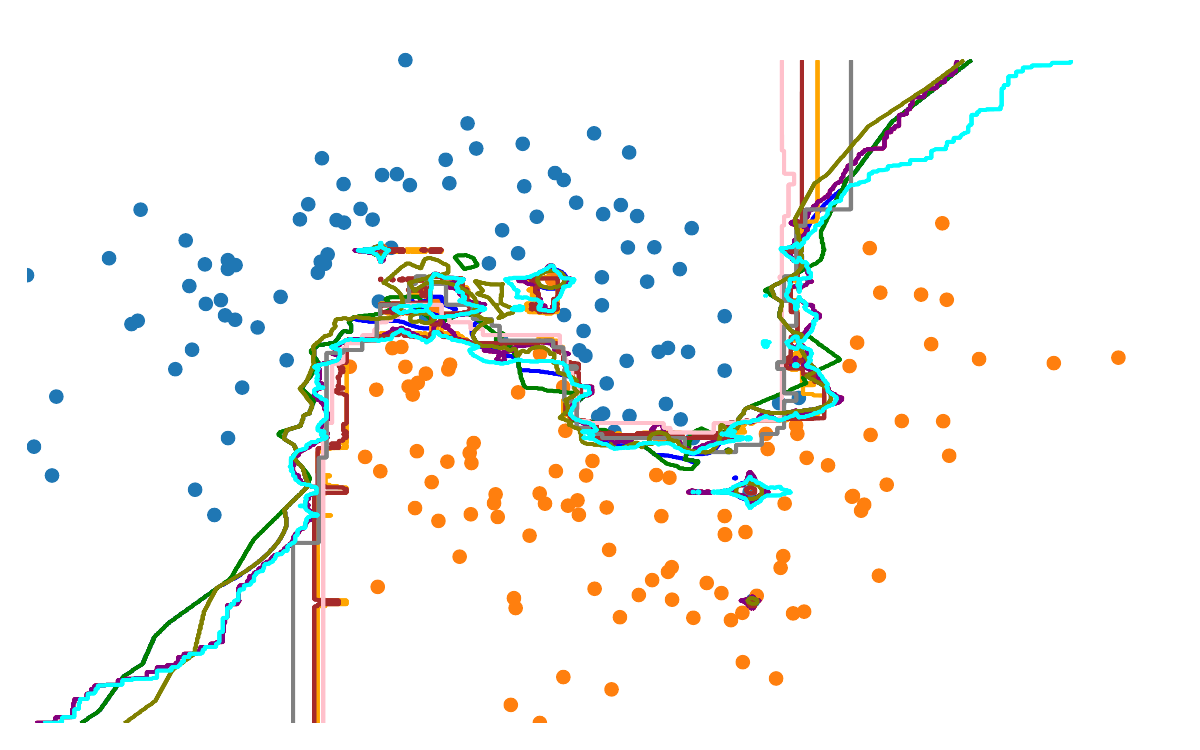}
    \caption{Illustration of prediction discrepancies of a pool of 10 models trained over the half-moons dataset (dots) using Autogluon. Each colored line represents the decision boundary of a classifier.}
    \label{fig:illustration-halfmoons}
\end{figure}

The machine learning practice leverages a large variety of models and techniques to tackle classification tasks.
Optimizing all the parameters and hyper-parameters involved leads to an arbitrary large number of models that may end up achieving similar classification performances, while having learned significantly different classification patterns.
This phenomenon, which we call \textit{prediction discrepancies}, arises from the fact that any dataset is an imperfect sampling of a classification task and the performance evaluation is done over a non exhaustive validation set.

Prediction discrepancy questions the selection of one model over others with similar performances.
Indeed, this choice is often made blindly, since the differences between models and their limits are generally not observable. 
This is all the more problematic at a time of widespread development of machine learning-based applications, when society calls for an increased level of responsibility around the deployment of AI systems.
Prediction discrepancies may result in concrete negative consequences (e.g. loss of opportunity, fairness), and should therefore be addressed.

Discrepancies in models are not always seen as an issue: aggregating the diverse predictions of weak but diverse classifiers is for instance the principle of ensemble learning.
Nonetheless, more recently, undesired consequences have been formalized, such as the threat of \emph{fairwashing} and explanation manipulation~\cite{aivodji2019fairwashing,anders2020fairwashing,slack2020fooling} or vulnerability to explanation inconsistency (see e.g.~\citet{Pawelczyk2020,Barocas2020} in the context of counterfactual explanations, and~\citet{DongRu2020} for global feature importances).
Some recent studies directly address this phenomenon~\cite{semenova2019study,Marx2019,Geirhos2020,Pawelczyk2020}.
However, the only tools provided by the literature to assess prediction discrepancies are aggregated metrics to quantify the importance of the phenomenon.
This information is neither precise nor actionable enough for a machine learning practitioner to tackle the issue.

Yet, this phenomenon can hardly be ignored, as its prevalence, empirically shown in this paper, suggests. In this context, designing more practical solutions is all the more crucial. For this purpose, we propose DIG (\emph{Discrepancy Intervals Generation}), an algorithm to capture and explain discrepancies in models at prediction level.
It is built on top of a comprehensive formalization of the phenomena and associated challenges.
The approach captures effectively the discrepancies in models and returns actionable explanations of the discrepancies in the form of discrepancy intervals.
These explanations allow the practitioner to choose among model candidates trained on the same data and to take actions on instances being affected by discrepancies.
The method introduced is model-agnostic.
It proposes grounded explanations and efficient computations as it takes advantage of the training data to discover discrepancies areas in the feature space.
To the best of our knowledge, this method fills a gap in the literature, since contrary to other interpretability approaches, it focuses on explaining the differences between models trained on the same data.

Section~\ref{sec:context} is devoted to providing the necessary context and formalization, as well as empirical and theoretical motivations for this work.
A novel approach to capture and explain discrepancies is explained in Section~\ref{sec:proposition-explaining-discrepancies}.
The effectiveness of the proposed method is shown in Section~\ref{sec:evaluation}.
Finally, the output of the approach is shown in Section~\ref{sec:usecases}.

\section{The Issue of Prediction Discrepancies}
\label{sec:context}

The machine learning practice is able to produce good model candidates when it is given a dataset, an hypothesis space and a learning algorithm.
Data points from the dataset guide the learning algorithm to select the best possible model from the hypothesis space (\textit{i.e.} the set of all model candidates).
In this process, known pitfalls prevent in reality the selection of an ideal model that will perfectly generalize.
Examples of pitfalls include datasets that in practice are sparse and noisy descriptions of the considered phenomena, hypothesis spaces that may not contain a good model candidate, or learning algorithms that lead to sub-optimal model candidates.
Given these pitfalls, many model candidates are able to reach similar predictive performance scores.
To illustrate this point, Figure~\ref{fig:illustration-halfmoons} shows 10 different model candidates with equivalent predictive performance scores on a toy dataset.
However, these models are significantly different, as shown by their decision boundaries.
In the following, we show that this issue, which we call \textit{discrepancies in models}, is prominent in the machine learning practice leading to negative consequences for model users.
In order to carry out this study, we first define the problem and the notations that will be leveraged along the paper.

\subsection{Problem Definition}

We consider a typical classification task defined on a feature space~$\mathcal{X} \subseteq \mathbb{R}^n$ associated with a target space $\mathcal{Y} = \{-1, 1\}$.
Binary classification is considered for the sake of clarity, but extension to the general case is straightforward.
The objective is to learn the best model for the classification task $f:\mathcal{X}\rightarrow\mathcal{Y}$ leveraging a dataset $\mathbb{X}=(\boldsymbol{X},\boldsymbol{y})$ sampled from the feature space.
The dataset is split into a training set $\mathbb{X}_{train}$ and a validation set $\mathbb{X}_{val}$. 
$\mathbb{X}_{train}$ is used to learn model candidates and the empirical risk of~$f$ denoted $\mathcal{\hat{R}}(f, X_{val})$ is evaluated on~$\mathbb{X}_{val}$ through common error functions (accuracy, F1 score...).
First, to simulate and observe \textit{discrepancies in models}, we introduce the \textit{pool} of models that gathers model candidates with the same predictive performance score on a dataset.

\begin{definition}[equi-performing pool]
Given the dataset~$\mathbb{X}$, we call the set of classifiers ${\mathbb{F}_{\mathbb{X}_{val}, \mathcal{\hat{R}}}=\{f_i\}_i}$ trained over $\mathbb{X}_{train}$ an \emph{equi-performing pool} of classifiers over $\mathbb{X}_{val}$ if all classifiers of $\mathbb{F}_{\mathbb{X}_{val}, \mathcal{\hat{R}}}$ have identical classification performance over $\mathbb{X}_{val}$.
That is to say: 
$$\forall f_i,f_j \in \mathbb{F}_{\mathbb{X}_{val}, \mathcal{\hat{R}}}, \mathcal{\hat{R}}(f_i, \mathbb{X}_{val}) = \mathcal{\hat{R}}(f_j, \mathbb{X}_{val})$$
\label{def:pool_strict}
\end{definition}

We denote without ambiguity $\mathbb{F}_{\mathbb{X}_{val},\mathcal{\hat{R}}}$ as $\mathbb{F}$.
This definition of a pool is very strict and doesn't account for models that are marginally different in predictive performance.
We relax this definition to allow similar, albeit different performances for the classifiers in the pool:

\begin{definition}[$\epsilon$-comparable pool]
Given $\mathbb{X}$ and $\epsilon \geq  0$, we call the set of classifiers $\mathbb{F}_\epsilon~=~\{f_i\}_i$ trained over $\mathbb{X}_{train}$ an \emph{$\epsilon$-comparable pool} of classifiers if each classifier of $\mathbb{F}_{\epsilon}$ has similar classification performance over $\mathbb{X}_{val}$ at an $\epsilon$ level.
That is to say: 
$$\forall f_i,f_j \in \mathbb{F}_\epsilon, \vert\hat{\mathcal{R}}(f_i, \mathbb{X}_{val}) - \mathcal{\hat{R}}(f_j, \mathbb{X}_{val})\vert \leq \epsilon$$
\label{def:pool_comparable}
\end{definition} 

To illustrate this definition, the 10 model candidates plotted Figure~\ref{fig:illustration-halfmoons} have less than $\epsilon=5\%$ of difference in accuracy.

This definition effectively means that the worst performing classifier of the pool should have a validation error of at most $\mathcal{\hat{R}}(f_{best}, \mathbb{X}_{val}) + \epsilon$, with $f_{best}$ the best performing classifier of $\mathbb{F}$.
Hence, a pool of equi-performing classifiers is a pool of $\epsilon$-comparable classifiers when no difference in prediction is tolerated: when setting $\epsilon=0$, we have: ${\mathbb{F}=\mathbb{F}_{\epsilon}}$.
This definition is similar to the notion of \emph{$\epsilon$-Level Set} introduced by~\cite{Marx2019}, except here the empirical risk is not evaluated on the training set.

The definition of pool discrepancy, the subspace where discrepancy in models are located, is then:

\begin{definition}[Pool discrepancy]
Given an $\epsilon$-comparable pool of classifiers~$\mathbb{F}_{\epsilon}$ trained over~$\mathbb{X}_{train}$, we call the \emph{pool discrepancy} the subspace $\mathcal{D}_{\mathbb{F}_{\epsilon}}\subseteq\mathcal{X}$ where at least one classifier from $\mathbb{F}_{\epsilon}$ disagrees in prediction with the rest of the pool. 
That is:
$$
\mathcal{D}_{\mathbb{F}_{\epsilon}} = \{ x \in \mathcal{X} \quad / \quad \exists f_i,f_j \in \mathbb{F}_\epsilon \text{ s.t. } f_i(x) \neq f_j(x) \}
$$
\end{definition}

On Figure~\ref{fig:illustration-halfmoons}, the pool discrepancy is located between the decision boundaries of the different models, where instance predictions change depending on the choice of the model.

$\mathcal{D}_{\mathbb{F}_{\epsilon}}$ is not a \emph{set} of existing instances for which classifiers disagree, but the (not-necessarily connected) \emph{region} of the feature space~$\mathcal{X}$ in which sampled instances are attributed conflicting predictions.
In the event where all classifiers of~$\mathbb{F}_{\epsilon}$ have strictly identical behaviors over the whole feature space, $\mathcal{D}_{\mathbb{F}_{\epsilon}} = \emptyset$.
However, in this work we argue that in reality even though classifiers have similar performances, the pool discrepancy~$\mathcal{D}_{\mathbb{F}_{\epsilon}}$ can be significantly large.

\subsection{Prediction Discrepancies are Inherent to the Machine Learning Practice: an Experiment}

Pool discrepancy is an information hidden by the performance scores commonly used by machine learning practitioners.
Indeed, these scores are aggregated statistics over the whole dataset: two models can make  prediction errors in different locations of the feature space while still exhibiting the same score.
But how prominent is the phenomenon in practice?
If we replicate the behavior of machine learning practitioners designing a model, how many prediction discrepancies do we expect?
How does the phenomenon of prediction discrepancies generalizes across datasets?
To answer those questions, we perform the following experiment.

To replicate machine learning practitioners and use credible model candidates on a dataset, we leverage OpenML~\cite{OpenML2013}.
OpenML hosts curated datasets concurrently with trained model candidates developed and submitted by machine learning practitioners.
It also gives access to the predictions made by these models.
The experiment is performed on the OpenML classification benchmarking suite, OpenML-CC18~\cite{oml-benchmarking-suites}, which gathers 72 datasets of all sizes with realistic complexity on various problem domains.
To create an $\epsilon$-comparable pool of classifiers with $\epsilon=2\%$, we retrieve the best 100 model candidates that are \textit{at worse} $2\%$ less accurate than the best performer of the dataset.

\begin{figure}[h]
    \centering
    \includegraphics[width=0.7\linewidth]{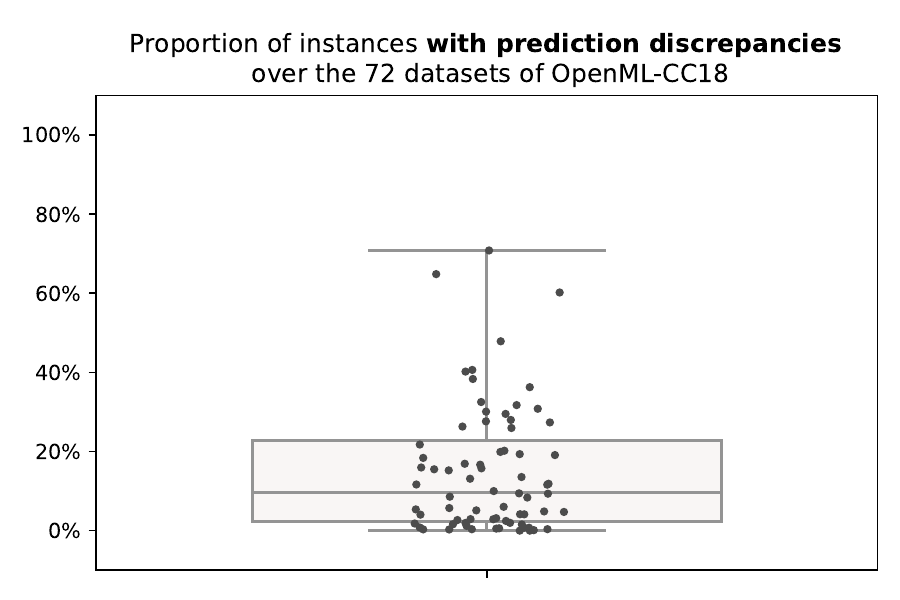}
    \caption{Proportion of prediction discrepancies among the 72 datasets of OpenML-CC18 benchmark suite. Each dot is a dataset where the ordinate is the proportion of instances affected by prediction discrepancies, among the best models submitted on OpenML by ML practitioners (in a $2\%$-comparable pool of classifiers).}
\label{fig:proportion_discrepancies}
\end{figure}

The results are shown in  Figure~\ref{fig:proportion_discrepancies}, in which we observe that 70 datasets out of the 72 of OpenML-CC18 present prediction discrepancies among the best model candidates submitted by machine learning practitioners.
The 2 remaining ones present singular characteristics: 
one has classifiers with perfect accuracy, and the pool of the other consists of a single model, as the others are below the $2\%$ threshold in accuracy.
For half of the 72 datasets, 10\% of their instances have prediction discrepancies and 20 datasets among the 72 have more than 20\% of prediction discrepancies.
The full results are available in Appendix~\ref{tab:app-survey}), all the predictions are available on OpenML and attached to this paper the code to replicate the experiment.

Thus, a very significant proportion of instances of the extensive OpenML-CC18 benchmark have prediction discrepancies, to an extent that can't be explained solely by the very small differences in prediction accuracies of the models in the pools (i.e. less than 2\%).
Additionally, we conduct an analysis of the distribution of model discrepancies.
The experiment, shown in Appendix~\ref{appendix:wasserstein}, suggests that instances with prediction discrepancies are rather grouped and similar in the feature space, and that a significant source of pool discrepancy may be instances of opposite classes that are too close for models to be able to properly separate them. 

\subsection{Addressing Prediction Discrepancies}


From the observations made in the previous section, one can thus expect from real world applications to also present significant prediction discrepancies.
Yet, when prediction discrepancies go undetected by model developers, negative consequences can arise.
For instance, a credit applicant may be arbitrarily accepted or rejected if he/she is subject to discrepancy, depending on the selected model.

Recent literature has been viewing notions similar to prediction discrepancies as a potential threat~\cite{aivodji2019fairwashing,Damour2020,slack2020fooling,Pawelczyk2020,rawal2020CFdistributionshift}. Among these, works on
\emph{fairwashing}~\cite{aivodji2019fairwashing,anders2020fairwashing} and \emph{explanation manipulation}~\cite{dombrowski2019explanations,slack2020fooling} have shown that prediction discrepancy can be maliciously leveraged to generate models with seemingly acceptable predictive performance, but completely manipulated explanations, i.e. potentially hiding harmful biases. On a different note, \citet{Pawelczyk2020} highlight the risk of inconsistent behavior for counterfactual explanations when there is discrepancy.

However, most of these works merely acknowledge the issue, either to avoid it or highlighting associated risks, without trying to understand it directly.
A notable exception and the most related work to ours is \citet{Marx2019}, which calls the phenomenon \emph{predictive multiplicity} and proposes a study of the potential threats it poses.
In order to address the issue of prediction discrepancies, the authors proposed several numerical criteria to measure the discrepancy of a pool consisting in the class of $\epsilon$-performing (a notion similar to the one of \textit{equi-performing pool} we introduced previously, but with the performance being evaluated over the training set) linear models, a focus that is also made in the cases of~\citet{semenova2019study,DongRu2020} and~\citet{krco2023mitigating} in the context of fairness.
\citet{Marx2019} thus show that competing models may have significantly different behaviors without being reflected on the empirical risk $\mathcal{\hat{R}}$.
\citet{Geirhos2020} conducts a study on prediction agreement using Cohen's Kappa coefficient~\cite{Cohen1960}.
This coefficient allows to compare models with different accuracies by normalizing the prediction agreement by the accuracy of the classifiers.
More broadly, the phenomenon of discrepancy in models can be related to the work of \citet{Damour2020} about the under-specification of machine learning tasks.


Apart from \citet{zeev2018controversy}, which uses pattern mining algorithms to identify data subgroups with high disagreement, and to the best of our knowledge, existing works focus on the quantification of discrepancies in models globally, at domain scale $\mathcal{X}$ (e.g. ambiguity and discrepancy scores in the case of~\cite{Marx2019}).
Such global scores help to raise awareness about the issue.
But they cannot help the machine learning practitioner addressing it by \emph{understanding where in the feature space the discrepancies in models occur}.

In the following, we propose to address this issue by providing the practitioner with explanations for the prediction discrepancies, that would allow him/her to identify the possible actions to mitigate these issues.
Possible use-cases encompass (1) model debugging, to identify uncertain areas of the feature space to improve the modeling (e.g. through the labelling of more training instances), (2) remedial measures, such as abstention or human intervention, especially for high-stakes applications when instances are subject to prediction discrepancies or (3) certification or model auditing, to provide guarantees about discrepancies in models.

\section{DIG: Capture and Explain Discrepancies}
\label{sec:proposition-explaining-discrepancies}
\begin{figure}[t]
  \begin{center}
    \subfloat[Initial state: classifiers are trained on the training set.]{
      \includegraphics[width=0.27\linewidth]{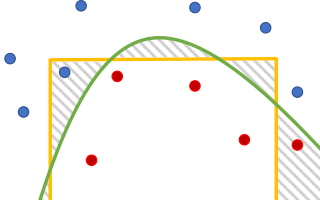}
      \label{sub:discrepancies_area}
        }
    \hspace{0.15cm}
    \subfloat[Step 1.1: instantiation of the graph ($k=3$): construction of the counterfactual segments]{
      \includegraphics[width=0.27\linewidth]{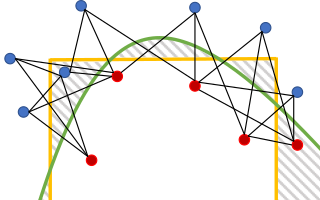}
      \label{sub:graph}
                         }
    \hspace{0.15cm}
    \subfloat[Step 1.2: refinement of one edge ($n_{epochs}=4$)]{
      \includegraphics[width=0.27\linewidth]{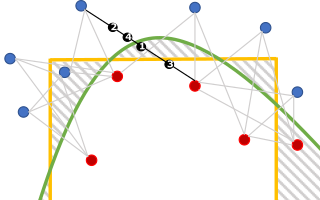}
      \label{sub:refinement}
                         }
     \hspace{0.15cm}
    \subfloat[Step 1.3: extraction of one discrepancy interval]{
      \includegraphics[width=0.27\linewidth]{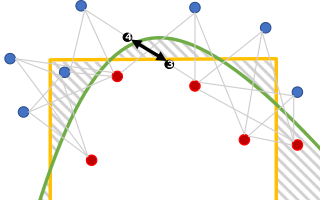}
      \label{sub:extraction_interval}
                         }
    \hspace{0.15cm}
    \subfloat[Step 1.3: all the discrepancy intervals]{
      \includegraphics[width=0.27\linewidth]{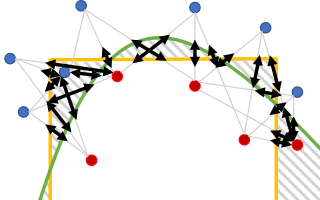}
      \label{sub:all_intervals}
                         }
    \hspace{0.15cm}
    \subfloat[Step 2: retrieval of the relevant discrepancy interval for a new instance]{
      \includegraphics[width=0.27\linewidth]{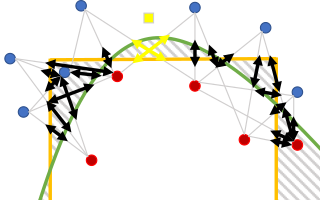}
      \label{sub:retrieval_intervals}
                         }
    \caption{Illustration of the principle of DIG on a toy dataset (red and blue points, colored depending on their true label) and a pool of 2 classifiers (yellow/green lines). Discrepancy regions to be detected by DIG are represented by hatched areas.
    }
    \label{fig:algo}
  \end{center}
\end{figure}


The literature in machine learning explainability focuses mainly on explaining either a model's prediction or an entire model \cite{guidotti2018survey}.
To our knowledge, the topic of explaining differences between models is not addressed.
We propose an algorithm to capture and explain discrepancies in models at prediction level.
It is model-agnostic, a paradigm more suited to the common machine learning practice, often relying complex predictive pipelines (with pre-processing, models and post-treatment...).
In the following, we explain the desired specifications for the problem and we describe the algorithm.

\subsection{Algorithm Specifications}
\label{sec:proposition-requirements}

We propose to capture the prediction discrepancies of a trained pool of classifiers, and explain, for a given prediction, the relevant discrepancy area (\textit{i.e.} locally in the feature space). Concretely, we formulate the following specifications to optimize the approach's relevance on real use-cases.


\begin{description}[leftmargin=0pt]

    \item[Requirement 1: Model-agnostic approach] In the ML practice, a predictive pipeline is usually complex with data pre-processing, model selection and post-treatment. To grant more flexibility to the user, the approach should be model-agnostic with no assumed access to the model apart from the input and output.
    \item[Requirement 2: Grounded \& Actionable explanations] 
    Many regions of the feature space are not densely covered by ground-truth data. Besides, all forms of explanations may not be adequate to gain insights about discrepancies. The method should provide grounded and actionable explanations.
    \item[Requirement 3: Precision of the explanations] To be useful, the explanations provided should faithfully describe the actual local discrepancies for a prediction.
    \item[Requirement 4: Efficient detection and explanation generation] The complexity needed to explore without knowledge the feature space $\mathcal{X} \subseteq \mathbb{R}^n$ is $O(2^n)$. To be usable, the proposed approach should rely on an efficient algorithm to explore the feature space. Moreover, to ensure a practical usage, the approach should be able to generate discrepancy explanations at prediction time.
\end{description}


Defining these requirements helps clarifying why existing explainability methods, although seemingly related, cannot be used to address the objective considered in this paper. Indeed, despite traditional model-agnostic local explainers such as LIME~\cite{ribeiro2016should} or SHAP~\cite{lundberg2017unified} also providing insights about the local decision boundaries of trained classifiers, they can not be directly leveraged to understand precisely where discrepancy regions are (\emph{Requirement 3: Precision of the explanations}), as they merely give an indication on the general direction of the decision boundary. Furthermore, they have been shown to suffer from various issues, such as a vulnerability to correlated features~\cite{kumar2020problems}, problems of locality~\cite{laugel2018defining} or general lack of trustworthiness~\cite{Rudin2019,laugel2019issues,slack2020fooling}, making their returned explanations possibly unreliable and therefore unable to satisfy \emph{Requirement 2: Grounded and Actionable explanations}.
Finally, the local post-hoc paradigm they consider, similarly to counterfactual explanations~\cite{wachter2017} or anchors~\cite{ribeiro2018anchors}, make them generally inefficient, hence unfit for Requirement~4 (\emph{Efficient detection and explanation generation}). 

Based on these conclusions, we therefore introduce a new algorithm, named DIG (Discrepancy Interval Generation).
DIG generates \emph{intervals of discrepancies}, explanations to describe the discrepancy areas relevant for a prediction.
These intervals are searched along \emph{counterfactual segments}, i.e. segments between points of the training set (See Figure~\ref{fig:algo}). Formulating the discrepancy explanations as counterfactual explanations~\cite{wachter2017} helps to satisfy the actionability requirement (Req. 2, \emph{Grounded and Actionable explanations}), as it allows to clearly delimit the areas where predictions change.
An example of the expected discrepancy explanation in a fictional 2-d case is:

\begin{displayquote}
"A female customer with age 38 has her loan application automatically denied by the model deployed in production by a bank. However, the data scientist responsible for this model should be careful, because $\epsilon$-comparable classifiers are \textbf{disagreeing} for this prediction. The disagreement \textbf{ranges for women between ages 37 and 40}."
\end{displayquote}


\subsection{Algorithm Description}
\label{sec:proposition-algorithm-description}

DIG takes as input a pool of classifiers $\mathbb{F}_\epsilon$ and the training set $\mathbb{X}_{train}$ to output explanations of the relevant discrepancies in the pool $\mathbb{F}_\epsilon$ for any instance $x \in \mathcal{X}$. Besides, the algorithm returns the information of whether a new instance lies in a discrepancy area or not, an information provided by the indicator function~$\mathbbm{1}_{\mathcal{D}_{\mathbb{F}_{\epsilon}}}(x)$. Given two training points $x_i,x_j \in \mathbb{X}_{train}$ defining a segment $[x_i,x_j]$, when this segment crosses an area of discrepancies, we call  \emph{interval of discrepancies} the portion of this segment that lies in the discrepancy area $[x_i,x_j]\cap \mathcal{D}_{\mathbb{F}}$.
This sub-segment is noted $[x_{k_1},x_{k_2}]$, with $x_{k_1}$ and $x_{k_2}$ the borders of the area of discrepancy on the segment $[x_i,x_j]$.

The proposed approach is divided in 2 steps: (1) a \textit{Learning step} to pre-compute the intervals of discrepancies from the pool between points of the training set (pseudo code in Algorithm~\ref{alg:pool2graph-preprocessing} and illustration in the first 5 figures of Fig.~\ref{fig:algo}) and (2) an \textit{Explanation Inference step} to ensure a fast generation of explanations for any new point, by approximating the relevant discrepancies (pseudo code in Algorithm~\ref{alg:pool2graph-inference} and illustration in the last figure of Fig.~\ref{fig:algo}).

\subsubsection{(Step 1) Learning: Capture Discrepancies over the Dataset}
\label{sec:proposition-step1}

\begin{algorithm}[h]
\caption{DIG (Step 1) Learning: capture discrepancies over the dataset}
\label{alg:pool2graph-preprocessing}
\begin{algorithmic}[1]
    \State {\bfseries Input:} Pool $F_\epsilon$, training set $(\mathbb{X}_{train}, y_{train})$, $k$, $n_{epochs}$
    
    \State Initialize graph $\mathcal{G}$
    \State For every instance $x \in \mathbb{X}_{train}$, create a node $v \in \mathcal{G}$ 
    \State Label each node $v$ with $F_\epsilon(x)$ and $\mathbbm{1}_{\mathcal{D}_\mathbb{F_\epsilon}}(x)$
    \State For every note $v_i$, create an edge $e_{i,j_k}$ between the $k$ nearest nodes to $v_i$ verifying Eq~\ref{eq:connection-condition}, noted $v_{j_k}$ 
    \For{$n$ in $n_{epochs}$}
        \For{every edges $e_{i,j} \in \mathcal{G}$ that comply with Eq~\ref{eq:connection-condition}}
            \State Create a new node $v_k$ associated with a new point $x_k = \frac{x_i+x_j}{2}$
            \State Label the new nodes with $F_\epsilon(x)$ and~$\mathbbm{1}_{\mathcal{D}_\mathbb{F_\epsilon}}(x)$
            \State Remove edge $e_{i,j}$ from $\mathcal{G}$ and create edges $e_{i,k}$ (resp. $e_{k,j}$) between $v_i$ and $v_k$ (resp. $v_k$ and $v_j$).
        \EndFor
    \EndFor
    
    \State {\bfseries Output:} Refined graph $\mathcal{G}$ 
    
\end{algorithmic}
\end{algorithm}

The goal of this first step is to produce \textit{intervals of discrepancies} as a mean to approximate the discrepancy regions~$\mathcal{D}_{\mathbb{F}_\epsilon}$ of an $\epsilon$-comparable pool. To ensure grounded explanations (Req. 2, \emph{Grounded and Actionable explanations}), the algorithm relies on the training set $\mathbb{X}_{train}$ to search and capture the prediction discrepancies.
Intervals of discrepancies are sought between pairs of points $(x_i,x_j) \in \mathbb{X}_{train}$ on one of two conditions: (1) at least one of them is attributed conflicting predictions by the pool or (2) the points $(x_i,x_j)$ have unanimously different predicted labels. That is to say, the algorithm searches for intervals of discrepancies along $[x_i, x_j]$ if it verifies the condition (lines 2 to 5 of Algorithm~\ref{alg:pool2graph-preprocessing}):
\begin{equation}
    \begin{cases}
        \mathbbm{1}_{\mathcal{D}_\mathbb{F}}(x_i)+ \mathbbm{1}_{\mathcal{D}_\mathbb{F}}(x_j) \geq 1\\
        \text{Or: }\mathbbm{1}_{\mathcal{D}_\mathbb{F}}(x_i) + \mathbbm{1}_{\mathcal{D}_\mathbb{F}}(x_j) = 0 \wedge \exists f \in \mathbb{F}_\epsilon, f(x_i) \neq f(x_j)
        
    \end{cases}
    \label{eq:connection-condition}
\end{equation}

Discrepancy areas can theoretically very well exist between two instances being unanimously predicted to belong to the same class, a case not covered by the conditions of Eq.~\ref{eq:connection-condition}. However, guiding the exploration this way helps speeding up the approach by focusing on areas where discrepancy can be found with high certainty. 

The search of the discrepancies is supported by a graph~$\mathcal{G}$.
Each point of the training set ${x \in \mathbb{X}_{train}}$ is represented in the graph by a single node~$v_i$.
To speed up calculations and since we expect some intervals to be redundant, instead of linking every pair of training instances, each node~$v_i$ is only connected to its~$k$ closest neighbors~$v_j$ (Euclidean distance in the feature space~$\mathcal{X}$) verifying Eq.~\ref{eq:connection-condition} with a single undirected edge~$e_{i,j}$.
Each node~$v_i$ is then labelled with its associated  coordinates in~$\mathcal{X}$, its predictions made by every model of the pool~$\mathbb{F}_\epsilon$ and whether there is prediction discrepancy or not~($\mathbbm{1}_{\mathcal{D}_\mathbb{F}}(x_i)$).
This step of the algorithm is illustrated Figure~\ref{sub:graph}.
Once instantiated with this procedure, the graph is a coarse representation of the intervals of discrepancies between points of the training set.
To obtain a better approximation of the intervals of discrepancies, the graph is refined to search for the borders~$\{x_k\}_k$ of the discrepancies areas along the edges.

The following search is conducted (lines 6 to 12 of Alg.~\ref{alg:pool2graph-preprocessing}): every edge $e_{i,j}$ of the graph $\mathcal{G}$ that verifies Equation~\ref{eq:connection-condition} is split in half (See Figure~\ref{sub:refinement}).
For each split, a new node $v_k$ is created.
This node is associated with a new point in the feature space $x_k \in \mathcal{X}$, in the middle of the segment $[x_i,x_j]$ associated with the edge $e_{i,j}$, that is $x_k = \frac{x_i+x_j}{2}$.
The new point $x_k$ is labelled by models of the pool $\mathbb{F}_{\epsilon}$ and by the indicator function $\mathbbm{1}_{\mathcal{D}_\mathbb{F}}(x_k)$ for the presence of discrepancy.
The former edge~$e_{i,j}$ is removed from the graph~$\mathcal{G}$ and replaced by two new edges~$e_{i,k}$ and~$e_{k,j}$ (line 10 of Alg.~\ref{alg:pool2graph-preprocessing}).

This refinement process is repeated~$n_{epochs}$ times, a user-defined parameter.
With each iteration, the intervals of discrepancies along each initial edge~$e_{i,j}$ become more precise around the true area of discrepancies of the pool (Req. 3, \emph{Precision of the explanations}), as shown in Figure~\ref{sub:all_intervals}).
In the worst case scenario, when every edge is split at each iteration, the complexity of the graph growth is $n_{edges}*2^{n_{epochs}}$ where~$n_{edges}$ is related to~$k$ and~$m$ the number of instances in the training set.

\paragraph{Handling categorical attributes}
\label{sec:proposition-noncontinuous}
In the presence of non-continuous data in the input space, the dichotomous split of the segments segment $[x_i, x_j]$ previously described can not be performed. To preserve this heuristic intact, we propose to handle categorical attributes through data augmentation, 
to effectively transform the problem into a continuous-only higher dimensional one. Concretely, given a categorical attribute $A$ and two instances $x_i$ and $x_j$ taking respectively the values $x_i^{A}$ and $x_j^{A}$ along this attribute, the idea is to replace at the initial step the segment $[x_i, x_j]$ with two alternate versions, $[\hat{x}_i, \hat{x}_j]$ and $[\Tilde{x}_i, \Tilde{x}_j]$. These new instances verify: $\hat{x}_i^A = \hat{x}_j^A = x_i^{A}$ and $\Tilde{x}_i^A = \Tilde{x}_j^A = x_j^{A}$, while all other attributes remain the same. 
This step, to be performed only once at the beginning, leads to exploring the discrepancy along continuous features only, as all segments would now be taking constant values over categorical features.

Since this data augmentation is performed after the initial graph has been built, i.e. when each node is connected to its $k$ closest neighbors, the question of which distance to use arises. Besides using the Euclidean distance on one-hot encodings of the categorical attributes, another possibility is to use distances adapted for mixed-type data, such as Gower's distance~\cite{gower1971general}. Although the experiments shown in the paper are conducted with the one-hot encoding strategy, some results obtained using Gower's distance are shown in Appendix~\ref{appensix:results-categorical}.

The refined graph~$\mathcal{G}$, with pre-computed intervals of discrepancies, is used to generate explanations for a prediction.

\subsubsection{(Step 2) Explanation Inference: Fast Generation of Discrepancy Intervals}
\label{sec:proposition-step2}

\begin{algorithm}[h]
\caption{DIG (Step 2) Explanation inference: fast generation of discrepancy intervals for a prediction}
\label{alg:pool2graph-inference}
\begin{algorithmic}[1]
    \State{\bfseries Input:} Refined graph $\mathcal{G}$, instance $x$ to be explained, number $l\geq1$ of discrepancy intervals
    
    \For{every discrepancy interval $I=[v_{k_1},v_{k_2}]$ found in the graph $\mathcal{G}$}
        \For{every node $v_k \in [v_{k_1},v_{k_2}]$ generated during Step~1}
            \State Compute the Euclidean distance $d(x,v_k)$ between $x$ and $v_k$
        \EndFor
        \State Define $d(I,x) = \min_{v_k \in [v_{k_1},v_{k_2}]}{d(x,v_k)}$, the distance between $x$ and the discrepancy interval $I$
    \EndFor
    
    \State {\bfseries Output:} Return the $l$ closest discrepancy intervals to $x$ to explain discrepancies
    
\end{algorithmic}
\end{algorithm}

In this second step, outlined in Algorithm~\ref{alg:pool2graph-inference} the goal is to map at test time a new instance to the relevant pre-computed intervals of discrepancies in the graph. When a new prediction is made, two pieces of information are returned to the user.
First, the prediction is flagged if it is affected by prediction discrepancy using~$\mathbbm{1}_{\mathcal{D}_\mathbb{F_\epsilon}}(x_i)$. 
Second, the discrepancies impacting the neighbourhood of the prediction's instance are explained.
For this purpose, the algorithm returns the closest discrepancy intervals $[x_{k_1},x_{k_2}]$ (learned during Step 1) to the new instance (See Figure~\ref{sub:retrieval_intervals}).
To provide additional grounding, each discrepancy interval returned can be accompanied by the two training instances $x_i,x_j \in \mathbb{X}_{train}$ delimiting the edge the interval belongs to ; that is to say, the final explanation returned is $[x_i,x_{k_1},x_{k_2},x_j]$.
To retrieve the "closest intervals of discrepancies", the algorithm minimizes the distances between~$x$ and the nodes of the graph belonging discrepancy intervals, and returns the~$k$ closest ones.

\subsection{Extension: Adaptation to image classification} 
\label{sec:extension-digcv}
As it is mainly designed for tabular data, applying DIG to other types of data such as images requires some adaptation, as it raises two major questions. First, the defined sampling along "counterfactual directions" (i.e. straight lines) is not appropriate in a high dimensional setting, resulting in non-realistic examples being generated. Second, a related issue is that the feature-by-feature explanation in the input space proposed by DIG is not actionable in the case of images (pixels are not naturally understandable).
These issues have been generally raised when trying to adapt model-agnostic explainability methods to image classification models. For instance, LIME~\cite{ribeiro2016should} proposes to split the images into superpixels and use these as features, but this does not allow continuous transformations from one instance to another which is one benefit offered by DIG. Other methods have considered autoencoders~\cite{guidotti2019vae}, which we propose to use to extend DIG, as described in the next section.

\begin{figure}
    \centering
    \includegraphics[width=0.7\linewidth]{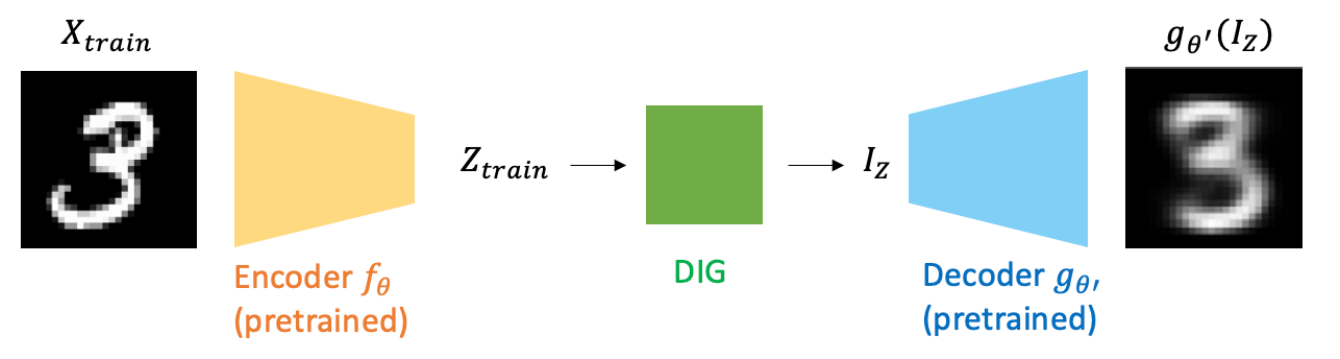}
    \caption{Proposed architecture for DIG-CV.}
    \label{fig:architecture}
\end{figure}

To circumvent these limitations, we propose to adapt DIG by adding a step of feature learning using variational autoencoders, more particularly a $\beta$-VAE~\cite{betavae}. This choice can be motivated by several reasons. First, the $\beta$-VAE loss function has been shown to favor the extraction of meaningful concepts. Second, transitioning from one image to the other by connecting in a straight line the representations learned by variational autoencoders in general has been already shown to generate meaningful images (see e.g.~\cite{kingma2013auto} for MNIST). 
The proposed architecture is represented in Figure~\ref{fig:architecture}. The exploration phase of the DIG algorithm is performed in the latent space~$Z$, with the assessment of discrepancy being performed on the reconstructions of these generated instances: $f_i(g_{\theta'}(I))$. We refer to this variant as DIG-CV.

\section{Evaluation}
\label{sec:evaluation}

Requirements~1 (\emph{Model-agnostic approach}, cf Section~\ref{sec:proposition-requirements}), 2 (\emph{Grounded \& Actionable explanations}) and 4 (\emph{Efficient explanation generation}) are intended to be guaranteed by design by DIG. Hence, in this section, we evaluate the last remaining Requirement 3 (\emph{Precision of the returned explanations}).
For this purpose, we address two questions.
First, we assess whether discrepancy areas are correctly detected during the Learning step of the proposed method.
Second, we evaluate the quality of the approximation of the local discrepancy areas by the returned discrepancy intervals.
 Finally, we test DIG againt two counterfactual explanation approaches, DICE~\cite{mothilal2020explaining} and Growing Spheres~\cite{laugel2018comparison}, to showcase its value.
The general framework for the experiments is described in Appendix~\ref{sec:appendix-experimental-protocol}.
The code required to fully replicate the experiments is made available\footnote{repository linked in first page}.

\subsection{Experiment 1: Does DIG Accurately Detect Discrepancy Areas?}
\label{sec:experiment-elephant}
The objective of this experiment is to assess that the proposed method effectively detects discrepancy areas in the feature space, i.e. that it generates nodes in discrepancy areas. As it is difficult to build a valid ground-truth to answer this question in an absolute manner, we rely on defining a baseline and compare its effectiveness to DIG's.

To this end, we use a Kernel Density Estimator~\cite{KDEparzen} (KDE) trained on $\mathbb{X}_{train}$. A Gaussian kernel is used, and the bandwidth is set to maximize the log-likelihood~\cite{habbema1974KDEselection}. Once fitted, the estimator is used to sample instances to explore the feature space and try to detect discrepancy areas. The number of instances generated is set to the number of nodes in the refined graph $\mathcal{G}$. Therefore, both sampling approaches (KDE and DIG) have the same budget in terms of number of generated instances, and the comparison between the two effectively focuses on \emph{where} these instances are sampled. Unless mentioned otherwise, the parameter values chosen for DIG are described in Appendix~\ref{sec:appendix-expe1}. 
Although only KDE  is used in this section, this experiment was also conducted with other commonly-used sampling approaches for tabular data, showing no particular difference in the obtained results: a Gaussian Mixture model~\cite{reynolds2009gaussian} and a Variational Autoencoder~\cite{kingma2013auto}. Full details on the implementation and the results for these other approaches are given in Appendix~\ref{appensix:results-expe1}.

\subsubsection{Illustrative Example}

\begin{figure}[t]
    \centering
    \includegraphics[width=0.44\linewidth]{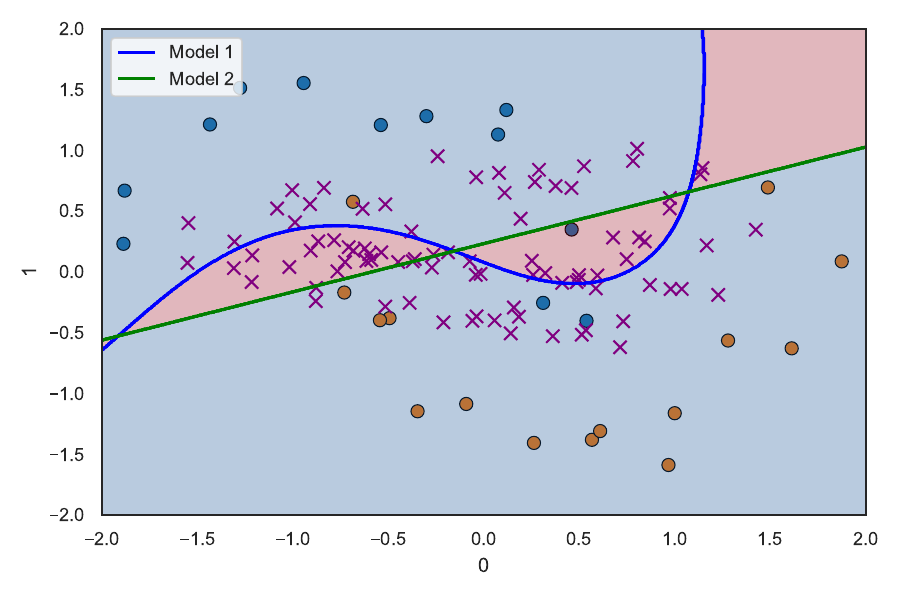}
    \includegraphics[width=0.44\linewidth]{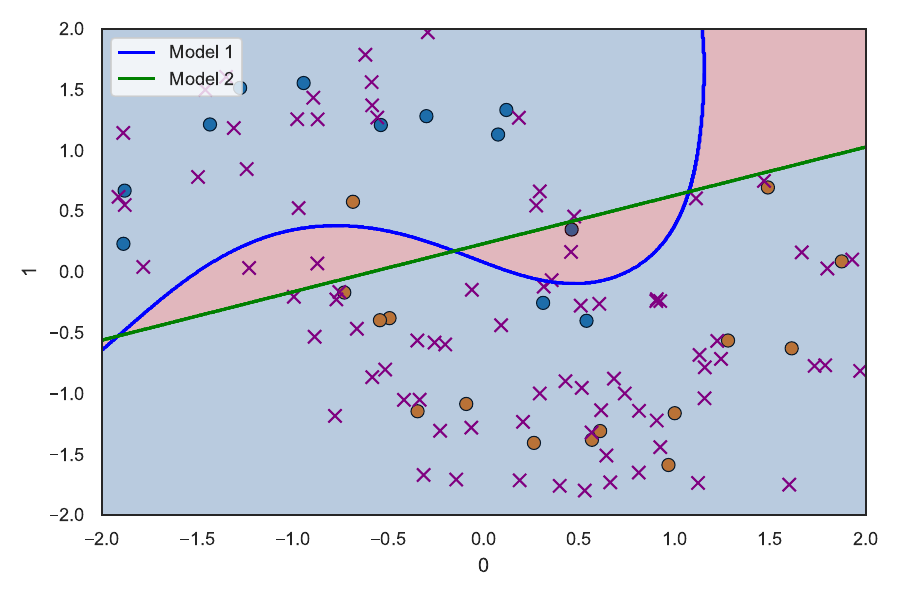}
    \includegraphics[width=0.44\linewidth]{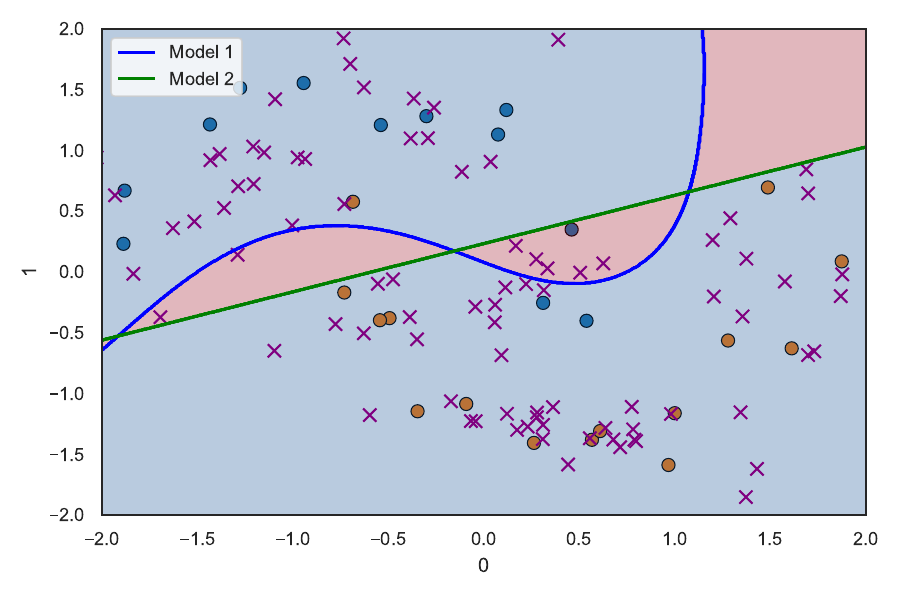}
    \includegraphics[width=0.44\linewidth]{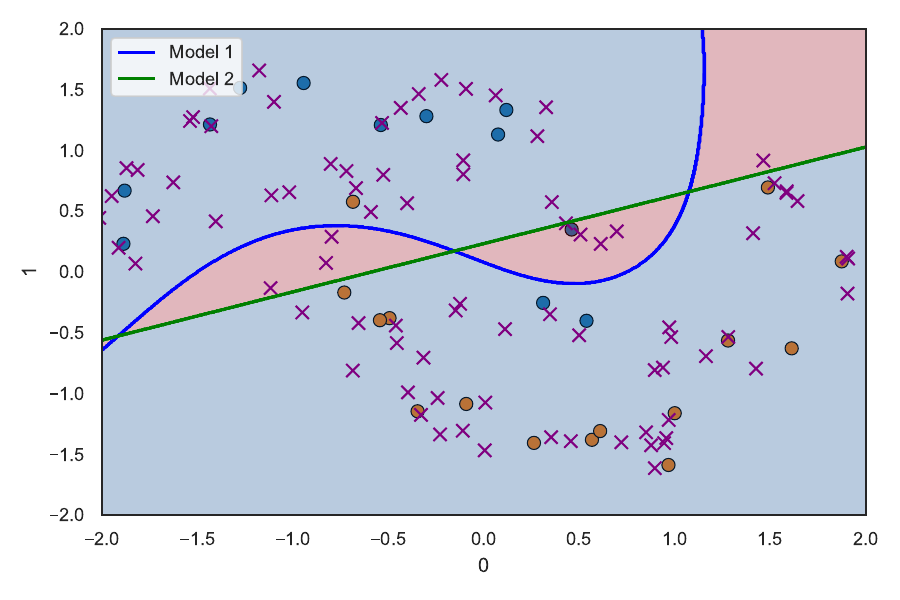}
    \caption{Sampling strategies of DIG (left) and KDE (right).}
    \label{fig:results-toyscenario-halfmoons}
\end{figure}

To illustrate their behaviors, examples of the two sampling approaches are shown in Figure~\ref{fig:results-toyscenario-halfmoons} for a 2-dimensional toy scenario. Two classifiers are trained on half-moons dataset (orange and blue dots represent the two classes): a logistic regression (decision boundary represented by the green line) and a SVM classifier with RBF kernel (blue line). We observe that large discrepancy regions can be found between the two lines.
Purple dots represent the instances generated by the two sampling strategies: the nodes~$\{v_i\}_i$ from the discrepancy graph of DIG (left figure), and the instances sampled by the KDE estimator (right figure).
The sampling of DIG is mainly localized on discrepancy areas, they are thus more accurately described.
On the contrary, the sampling of the KDE is not specifically localized on discrepancy areas, as expected.


\subsubsection{Black-box Scenario}
\label{sec:exeriment1-blackbox}

\begin{figure}[t]
    \centering

        
        
        
        
        
        

\begin{tabular}{l|c|c|c|c}
    Dataset & DIG & KDE & GMM & VAE \\
    \midrule
    half-moons & $\textbf{0.96\,(0.02)}$ & $0.92\,(0.03)$ & $0.94\,(0.02)$ & $0.92\,(0.02)$\\ 
    
    boston & $\textbf{0.78\,(0.05)}$ & $0.57\,(0.07)$ & $0.68\,(0.08)$& $0.45\,(0.03)$\\
    
    breast-cancer & $\textbf{0.75\,(0.05)}$ & $0.40\,(0.02)$ & $0.50\,(0.09)$& $0.55\,(0.08)$\\
    
    churn & $\textbf{0.60\,(0.02)}$ & $0.59\,(0.01)$ & $0.61\,(0.04)$& $0.60\,(0.01)$\\

    news & $\textbf{0.60\,(0.02)}$ & $0.42\,(0.05)$  & $0.50\,(0.05)$& $0.51\,(0.07)$\\
    
    adult & $\textbf{0.81\,(0.03)}$ & $0.60\,(0.02)$  & $0.62\,(0.03)$ & $0.62\,(0.04)$\\
    
    german & $\textbf{0.71\,(0.03)}$ & $0.65\,(0.02)$  & $0.60\,(0.03)$& $0.63\,(0.03)$ \\
    
    \bottomrule
\end{tabular}
    \caption{Results for the discrepancy detection evaluation.}
    \label{fig:results-evaluation-elephant}
\end{figure}

\begin{figure}[t]
    \centering
    \includegraphics[width=0.49\linewidth]{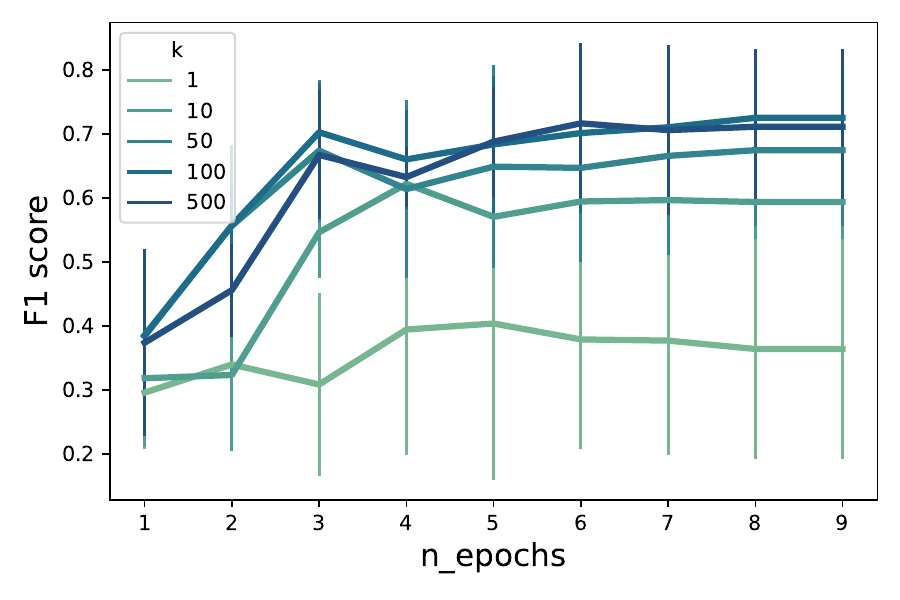}
    \includegraphics[width=0.49\linewidth]{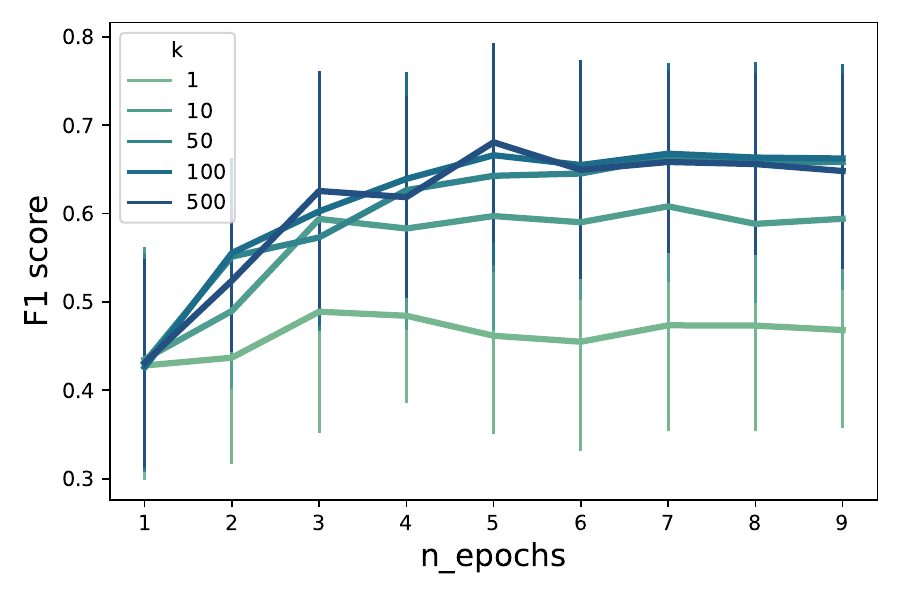}
        
    \caption{Average and standard deviation values of the F1 score obtained at each epoch for the Breast Cancer dataset (left) and the Boston dataset (right) for 10 runs.}
    \label{fig:results-elephant-curves-boston-breastcancer}
\end{figure}

We now propose to generalize this experiment to datasets with a higher dimension. Due to the novel nature of this work, we propose a new experimental protocol. In order to assess whether important discrepancy areas are efficiently detected by the two sampling strategies (DIG and KDE), we propose to measure to what extent the generated instances can be used to predict whether or not a new instance (drawn from $\mathbb{X}_{val}$) is going to be falling into an area of discrepancy. However, for the evaluation to be focused on the location of the sampling, the prediction model used on both strategies is set to be a 1-Nearest Neighbor classifier.
The intuition behind this evaluation is that if the graph adequately explores the feature space and detects discrepancies, then it should build nodes in these regions, and therefore, a Nearest Neighbor algorithm should be able to perform correctly.



Each instance~$x$ from the testing set~$\mathbb{X}_{val}$ is labelled with~$\mathbbm{1}_{\mathcal{D}_{\mathbb{F}_\epsilon}}(x)$.
The predictive performances of the approaches are then measured using the $F1$ score.
A score close to~$1$ indicates that discrepancy areas are well detected.

\paragraph{Results} Figure~\ref{fig:results-evaluation-elephant} presents the results of the experiment: the average and standard deviation values of the F1 scores obtained by DIG and the presented competitors over~10 runs.
Across the considered datasets, the accuracy of DIG varies between~$0.60$ and~$0.96$, and is often significantly higher than its competitor.
The exception to this rule is for the Churn dataset, where DIG achieves results comparable to KDE. The poor performance achieved by these approaches may be explained by the fact that it is highly imbalanced, making the task more difficult. 
These results thus tend to confirm that for the same budget, the exploration strategy of DIG seems to be capturing more effectively discrepancy areas compared to KDE. 

\begin{figure}[t]
    \centering
    \includegraphics[width=0.49\linewidth]{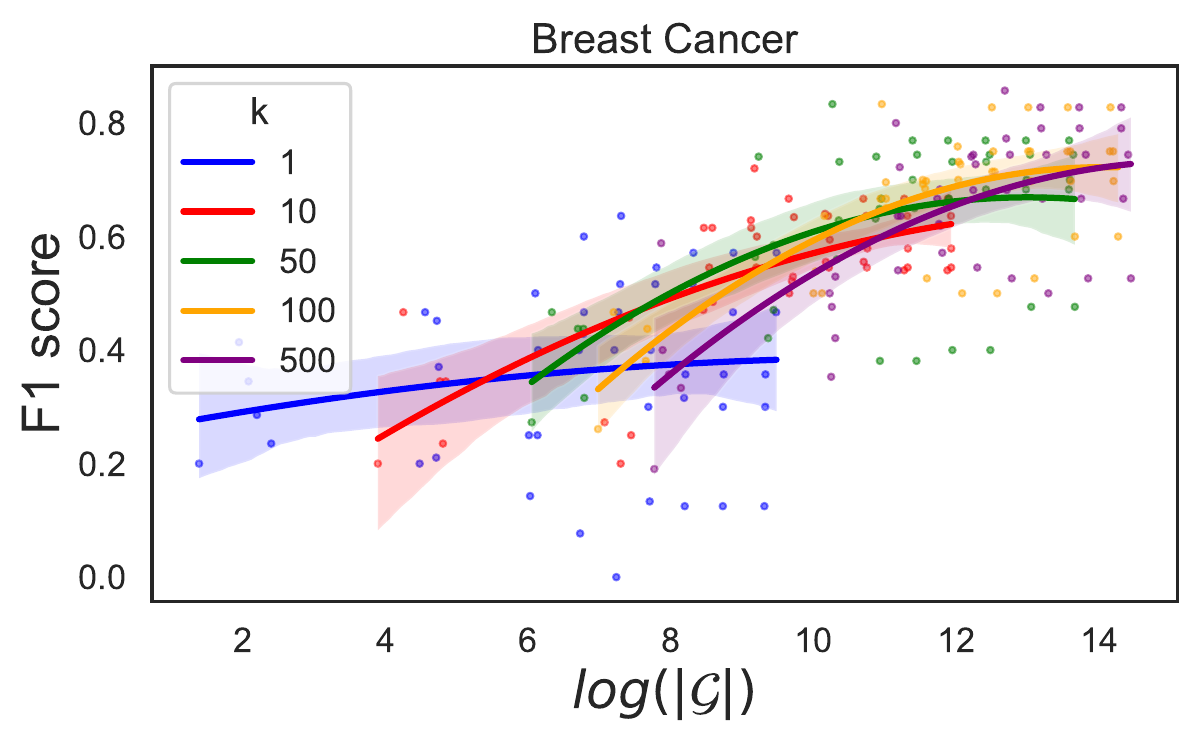}
    \includegraphics[width=0.49\linewidth]{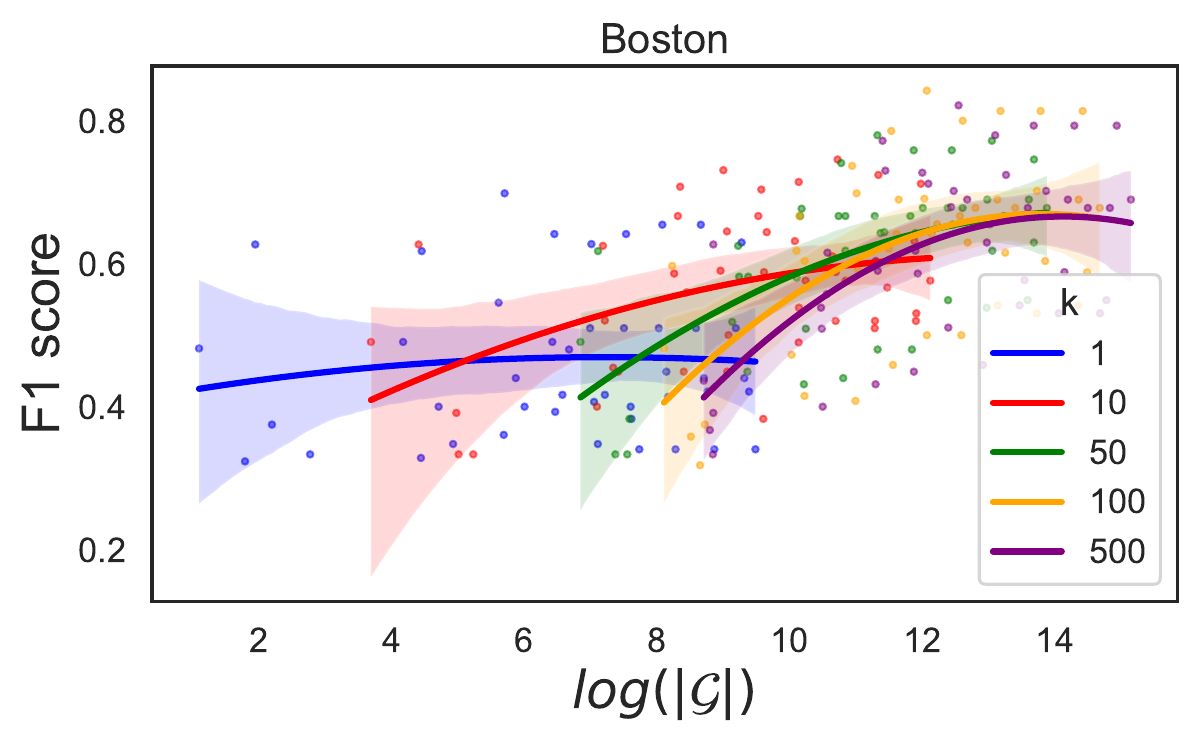}    
    \caption{Discrepancy detection accuracy (F1 score) depending on the number of nodes generated (X axis, measured by $log(\vert\mathcal{G}\vert)$), for various values of $k$ and $n_{epochs}$. Each color corresponds to a value of $k$, described in the label box; each dot to a certain value of $n_{epoch}$. Left: Breast Cancer dataset. Right: Boston dataset.}
    \label{fig:results-parameters-curves-boston-breastcancer}
\end{figure}

We assess the impact of the choice of the hyper-parameters~$k$ and~$n_{epochs}$ on the performances of DIG.
Figure~\ref{fig:results-elephant-curves-boston-breastcancer} presents the results for Breast Cancer and Boston datasets.
As expected, by allowing a higher number of instances to be sampled and therefore a better exploration, increasing the value of~$n_{epochs}$ ensures better performances of DIG. However, this effect becomes less significant for higher values, as important discrepancy areas already have been detected at this time.
A similar trend is observed with the values chosen for~$k$.
By definition, $k$ sets the numbers of discrepancy intervals that can be built, leading to more or less coverage of the feature space in terms of directions, a better exploration being expected as $k$ increases.
On the other hand, increasing~$n_{epochs}$ ensures that the directions already identified are more thoroughly explored with a better approximation of the discrepancy intervals, thus leading to a smaller increase in performances for higher values.
This can be observed as well Fig~\ref{fig:results-parameters-curves-boston-breastcancer}, which shows the F1 score obtained by DIG depending on the total budget, the number of nodes generated (in practice, we look at $log(\vert\mathcal{G}\vert)$), for various values of $k$ and $n\_epochs$.

Parameters $k$ and $n_{epoch}$ therefore both participate into controlling the exploration of the feature space performed by the proposed approach. Although in general, the higher these values are, the better the coverage is, this necessarily comes at the expanse of computation time. The optimal values to select thus depend on the dataset characteristics, as suggested by the ones selected to optimize the discrepancy detection (cf. Fig.~\ref{fig:results-evaluation-elephant}), and shown in Appendix~\ref{sec:appendix-expe1}.

\subsection{Experiment 2: Are Discrepancy Intervals Correctly Approximated?}
\label{sec:experiment-precision}

\begin{figure}[t]
    \centering
    \includegraphics[width=0.32\linewidth]{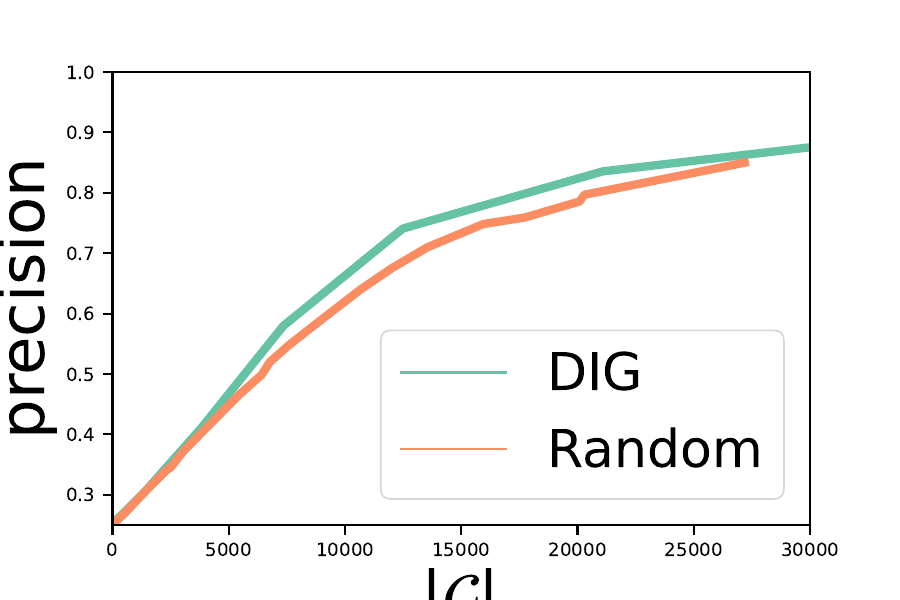}
    \includegraphics[width=0.32\linewidth]{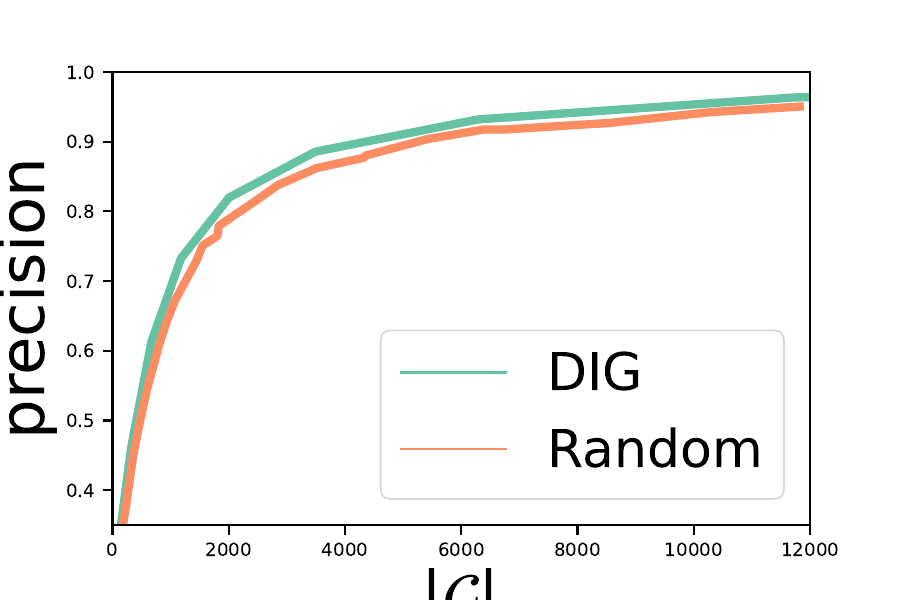}
    \includegraphics[width=0.32\linewidth]{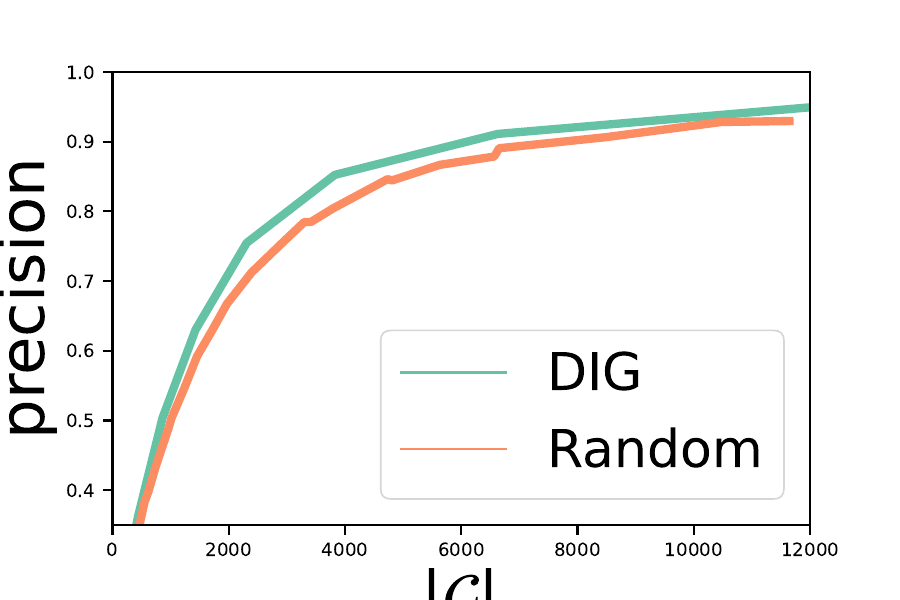}
    \caption{Comparison of the average discrepancy interval precision between DIG and a random split heuristic for the Adult (left), Boston (center) and Breast cancer (right) datasets.}
    \label{fig:results-evaluation-precisionvsrandom-curves}
\end{figure}

The objective of this second experiment is to assess the quality of the \textit{explanations} generated by DIG. More specifically, we evaluate if the the borders of the discrepancy intervals match the borders of the discrepancy area.


\paragraph{Evaluation criterion}
We calculate, for each refined discrepancy interval~$I_e=[x_{k_1}, x_{k_2}]$, the Jaccard index between $I_e$ and the actual local discrepancy area~$\mathcal{D}_e$. In practice, we perform a Monte-Carlo estimation of this value by drawing uniformly $100$ instances along $I_e$ that are labelled for discrepancy with the indicator function~$\mathbbm{1}_{\mathcal{D}_{\mathbb{F}_{\epsilon}}}(x)$ linked with the pool of classifiers.
This value is then averaged over all the generated intervals to obtain a \emph{precision} metrics. A value close to 1 indicates that discrepancy intervals accurately approximate the borders of the discrepancy areas.

\paragraph{Baseline}
Although seemingly intuitive, this evaluation of the precision is made difficult by the lack of proper baseline, let alone competitor.
We propose to build one by replacing the dichotomous split of the intervals performed at each epoch, with a random split of the segments. Concretely, when a segment $[x_i, x_j]$ is to be refined, instead of creating $x_k=\frac{x_i+x_j}{2}$, the competitor draws an instance $x_{k'}$ uniformly from $[x_i, x_j]$. 
In monotonous scenarios, binary search (dichotomic split) is expected to outperform a randomized uniform search. However, when looking for discrepancy areas, the uncertainty around the number of discrepancy regions makes this result non-trivial.

\paragraph{Results}
On Figure~\ref{fig:results-evaluation-precisionvsrandom-curves}, each sub-figure displays the average precision curves obtained with DIG and the Random Split competitor depending on the budget~$\vert\mathcal{G}\vert$, for three datasets. A first notable observation is the expected increase in precision over budget, illustrating the possibility of controlling the desired level of precision for the discrepancy estimation, obviously at the cost of computation time. This increase in precision follows a logarithmic curve, making the number of epochs not as worth over time and agreeing with the results observed in the previous section.
Moreover, these results show the superior efficiency of the dichotomic split along the discrepancy segments.

\subsection{Experiment 3: Comparing DIG with traditional XAI methods}

In this third experiment, we aim to illustrate the added value of DIG by showing that existing XAI methods cannot be easily and effectively used to understand discrepancy areas.

\paragraph{Competitors}
As discussed in Section~\ref{sec:proposition-requirements}, existing explainability approaches generally do not allow to explain discrepancy regions, and most (e.g. feature importance-based approaches such as LIME~\cite{ribeiro2016should}) cannot be easily adapted to tackle this problem. 
We therefore choose to focus on counterfactual explanation approaches~\cite{wachter2017,laugel2018comparison,mothilal2020explaining}, that we propose to adjust to our problem: instead of explaining the classification task, we aim to explain the discrepancy detection task, that we approximate by first training a global surrogate on the dataset $X$ with label $\mathbbm{1}_{\mathcal{D}}(x)$. 
We then build two competitors for DIG: one that performs counterfactual search at inference time (\emph{DICE}), and one that builds a set of discrepancy intervals in a preprocessing step (\emph{DiscGS}).
For the first approach, we leverage DICE~\cite{mothilal2020explaining}, a well-known counterfactual method that allows to explain a prediction by generating multiple diverse counterfactual examples. 
By generating two diverse counterfactual examples for a given instance~$x$, we thus define the borders of a local interval delimiting a discrepancy region around~$x$. We note $I_{DICE}(x)$ the local discrepancy interval generated for an instance $x$.

For the second approach, we build in a preprocessing step a set $\mathcal{S}$ of discrepancy intervals by generating, for each training instance, $2$ counterfactual examples that delimit the boundaries of the discrepancy regions.
For this purpose, we leverage \emph{GrowingSpheres}~\cite{laugel2018comparison}, a counterfactual method only requiring label access.
Given a test instance $x$ a local discrepancy interval is then retrieved among the set $\mathcal{S}$: $I_{DiscGS}(x) = \min_{I \in \mathcal{S}} {d(x, I)}$.

\paragraph{Evaluation criteria}
We aim to compare $I_{DICE}(x)$ and $I_{DiscGS}(x)$ to $I_{DIG}(x)$, generated using Algorithm~\ref{alg:pool2graph-inference}, along the following dimensions:
\begin{itemize}
    \item \textbf{Relevance of the interval for an observation $x$.}  As intervals are expected to describe the local discrepancy region around $x$, it is desired that the interval $I$ lies close to the observation $x$. We therefore measure $d(x, I) = \min_{a \in I} d(x,a)$, with~$d$ the Euclidean distance, using Monte-Carlo estimation.
    \item \textbf{Precision of the interval.} Intervals are expected to precisely delimit the discrepancy region, as stated by \emph{Requirement 3: Precision of the explanations}, defined in Section~\ref{sec:proposition-requirements}. We use the same precision criteria as in Experiment 2, that we note $p(I)$, measuring the overlap between the discrepancy region and the interval. A precision of $1.0$ means that the discrepancy region is very precisely delimited.
    \item \textbf{Time needed for inference.} To assess \emph{Requirement 4: Efficient detection and explanation generation}, we measure the time needed to compute all discrepancy intervals. Experiments were conducted on a 6-core Intel i7 16Go Mac computer. Only the time required for inference is evaluated: the global surrogate training for DICE as well as the graph building for DIG are not considered.
\end{itemize}

Similarly to previous experiments, for all datasets, we train a pool of equi-performing classifiers, and then generate local discrepancy intervals for all observations of the test sets. The average results are reported in Table~\ref{tab:expe3-results}.

\begin{table}[]
    \centering
    \resizebox{0.6\textwidth}{!}{\begin{tabular}{c||c|ccc}
        dataset & method & $d(x,I) \downarrow$ & $p(I) \uparrow$ & 
        time $\downarrow$\\
        \hline
        \multirow{3}{*}{half-moons}& DIG & $\textbf{0.05 (0.03)}$ & $\textbf{0.91 (0.11)}$ &
        $\textbf{21''}$\\
        &DiscDICE& $0.19 (0.31)$ & $0.63 (0.30)$ & 
        $22''$ \\
        &DiscGS& $\textbf{0.05 (0.04)}$ & $0.62 (0.35)$ &
        $22''$\\
        \hline
        \multirow{3}{*}{boston}& DIG& $\textbf{0.53 (0.24)}$ & $\textbf{0.97 (0.04)}$ &
        $13''$ \\
        &DICE& $1.40 (0.87)$ & $0.89 (0.20)$ & 
        $20''$\\
        &DiscGS& $0.72 (0.37)$ & $0.51 (0.33)$ & 
        $\textbf{12''}$\\
        \hline
        \multirow{3}{*}{german}& DIG& $\textbf{4.77 (1.56)}$ & $\textbf{0.95 (0.09)}$ & 
        $\textbf{2'33''}$ \\
        &DICE& $6.95 (1.97)$ & $0.79 (0.24)$ & 
        $2'54''$\\
        &DiscGS& $6.47 (1.55)$ & $0.76 (0.21)$ &
        $2'48''$\\
        \hline
        \multirow{3}{*}{breast-cancer}& DIG& $\textbf{0.85 (0.61)}$ & $\textbf{0.93 (0.05)}$ & 
        $\textbf{10''}$ \\
        &DiscDICE& $1.52 (0.95)$ & $0.92 (0.17)$ &
        $18''$  \\
        &DiscGS& $2.91 (1.32)$ & $0.68 (0.24)$ &
        $14''$\\
        \hline
        \multirow{3}{*}{churn}& DIG& $\textbf{0.30 (0.90)}$ & $\textbf{0.78 (0.40)}$ & 
        $\textbf{2'56''}$\\
        &DiscDICE& $1.41 (0.67)$ & $0.47 (0.31)$ &
        $3'28''$\\
        &DiscGS& $2.75 (0.52)$ & $0.35 (0.21)$ &
         $3'17''$\\
        \hline
        \multirow{3}{*}{adult}& DIG& $\textbf{0.10 (0.15)}$ & $\textbf{0.75 (0.25)}$ &
        $\textbf{4'31''}$\\
        &DiscDICE& $0.33 (0.28)$ & $0.51 (0.32)$ &
        $10'18''$  \\
        &DiscGS& $0.20 (0.21)$ & $0.43 (0.42)$ &
        $4'54''$\\
        
    \end{tabular}}
    \caption{Comparison results between DIG, DICE and DiscGS for various quality criteria. The arrows indicate whether is criteria is to be minimized (downward arrow) or maximized (upward).}
    \label{tab:expe3-results}
\end{table}

\paragraph{Results}
DIG achieves better results than DICE and DiscGS for all datasets and all the considered criteria, with the only exception being a slightly faster generation time on the Boston dataset. DIG thus generates discrepancy intervals that are more relevant for the considered predictions and more precise. 

In terms of inference time, DIG also proves to be more efficient than its counterparts. At inference time, this is especially true for DICE: generating local discrepancy intervals for the instances of the test set of the Adult dataset is $2.3$ times faster with DIG than with DICE. 
This is true at the preprocessing step as well (not shown in the table): it took $1'10''$ to build DIG's discrepancy graph for the German Credit dataset, against $44'49''$ for building DiscGS's set of intervals $\mathcal{S}$.
Although expected, as post-hoc methods typically rely on exploring the feature space through random generation and are therefore quite slow, this underlines the efficiency of DIG for the considered task.

These results therefore show that existing XAI methods are not well-suited to answer the studied problem of explaining discrepancy areas, since: (i) they require complex heuristic adaptations, and (ii) these adaptations are less satisfactory. On the other hand, DIG proves to be an efficient solution providing better results.  

\section{Explanations Generated by DIG: application to German Credit}
\label{sec:usecases}

In this section, we present a use-case on the German Credit dataset, illustrating the outputs of DIG and their usage in practice.
To showcase how the insights provided by the proposed method can be leveraged, we consider a scenario where a Data Scientist working in a bank would train a ML pipeline in order to estimate the default risk of customers. For this purpose, we leverage the German credit dataset.
In the following subsections, we use the methodology proposed in this paper to detect discrepancies over predictions in order to: (1) avoid taking actions with negative consequences based on this prediction (Sec.~\ref{sec:usecase-local}) (2) understand the sources of discrepancies and improve the modelling (Sec.~\ref{sec:usecase-global}.
In the following subsections, we use the methodology proposed in this paper to detect discrepancies over predictions and mitigate issues in two manners. First, we generate \emph{local} discrepancy explanations by extracting the most relevant discrepancy interval for a given prediction, allowing us to avoid taking actions with negative consequences based on this prediction (Sec.~\ref{sec:usecase-local}).
Second, we generate \emph{global} insights on discrepancy regions by aggregating local discrepancy explanations, allowing us to understand the sources of discrepancies and improve the modelling (Sec.~\ref{sec:usecase-global})).

Instead of using an automated machine learning pool like in previous sections, we compare two classifiers with similar performance, representing competitors among which the Data Scientist aims to choose a model to put in production: a SVM classifier (F1 score on test data of $0.85$) and an XGBoost classifier ($0.84$ on test data). To better understand the behaviors of these models, the Data Scientist uses DIG: graph~$G$ is refined following the methodology exposed in the previous sections. More details about the experimental protocol can be found in Appendix~\ref{appendix:usecases}.

\subsection{Local Discrepancy Understanding with Discrepancy Intervals}
\label{sec:usecase-local}
\begin{figure}[t]
  \begin{center}
      \includegraphics[width=0.8\linewidth]{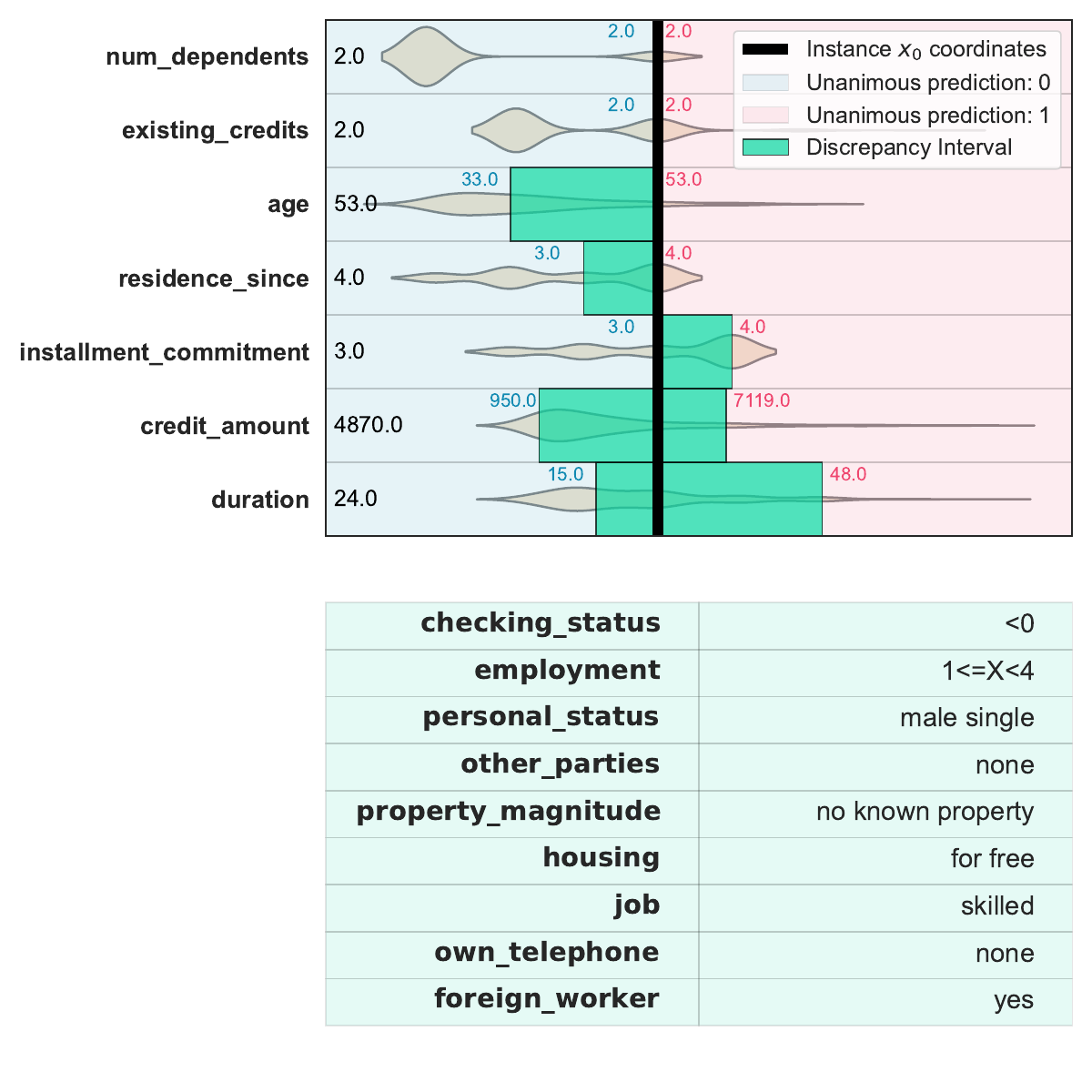}
    \caption{Output of DIG for an instance~$x_0$ of German Credit.}
    \label{fig:discrepancy_interval}
  \end{center}
\end{figure}

We illustrate the output of DIG with an explanation of the discrepancy areas discovered for an instance of the German credit dataset.
The explanation is displayed Figure~\ref{fig:discrepancy_interval}.
It is the closest discrepancy interval generated in~$\mathcal{G}$ to~$x_0$. As the input space consists of continuous and categorical features, the output contains two parts: an interval visualization for continuous features (top) and a table for categorical ones (bottom). 

\emph{Continuous features.}
For each continuous feature (rows), the visualization is centered around the coordinates of $x_0$ (black line, with coordinates written in black). The rest of the rows is then splitted between three colors: (i) blue, showing the feature range along the counterfactual direction of the interval where the pool uniformly agrees on predicting class~$0$ (ii) red, where the pool uniformly agrees on predicting class~$1$ (iii) green, where the pool disagrees. The green area along each feature is thus the refined discrepancy interval learned by DIG. The borders delimiting the discrepancy segment are written in blue and red. The explanation should be read like the following: \emph{models are disagreeing over their prediction for~$x_0$. Locally, they are disagreeing along the feature Age for people between~$33$ and~$53$ years old}.
To provide context and assist the interpretation of the explanation, the violin plots represent the distribution of the training data along each feature (in yellow).

\emph{Categorical features.}
The table shows the categorical attribute values shared by the discrepancy interval and~$x_0$. These attributes are shared by the two initial points defining the segment after the data augmentation step (Section~\ref{sec:proposition-step1}).


Besides understanding that the prediction of one model should not be trusted for this decision because models disagree over it, the model user is able to understand over the influence of which features the classifiers disagree locally. In face of the caveats identified by~\cite{Barocas2020}, this local insight also indicates that local explanation approaches for this prediction, e.g. a counterfactual one, would need to be especially cautious about certain features to avoid generating misleading explanations (e.g. explanations that would not hold after retraining).

\paragraph{Illustrative results for DIG-CV} Some examples of the outputs obtained with DIG-CV (Sec.~\ref{sec:extension-digcv}) for image classification on the MNIST and Fashion-MNIST datasets are given in Appendix~\ref{appendix:digcv}

\subsection{Models Behavior Investigation with Global Insights on Discrepancy}
\label{sec:usecase-global}

\begin{figure}[t]
    \centering
    \subfloat{
      \includegraphics[width=0.54\linewidth]{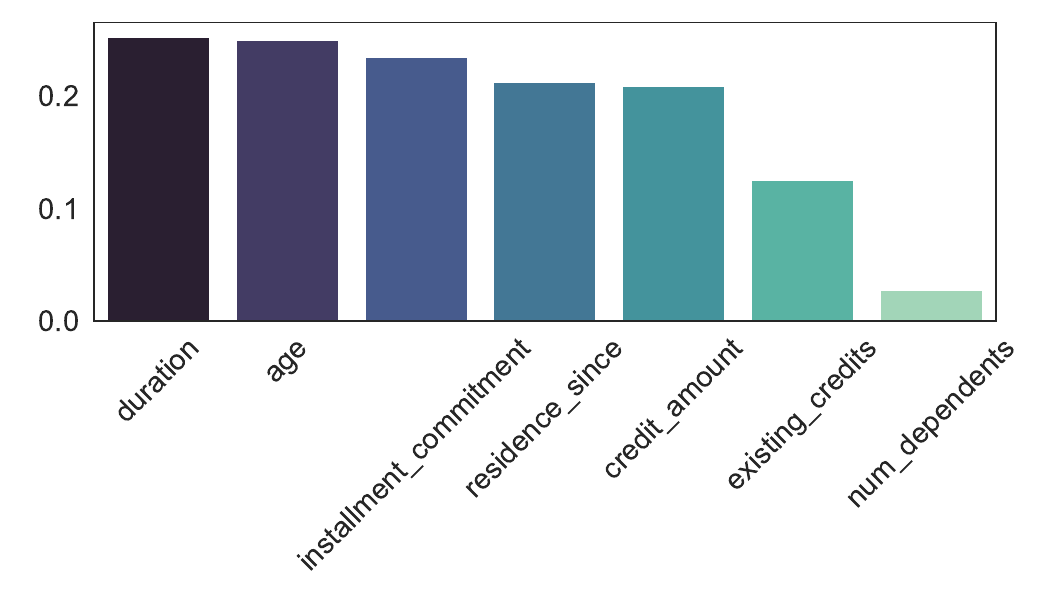}
    }\\
    \subfloat{
    \vspace{10pt}
    \resizebox{0.7\linewidth}{!}{
    \begin{tabular}{l|c|c|c|c}
        Segment description  & Disc. score & Expo & F1(SVM) & F1(XGB)\\
        \midrule
        \vtop{\hbox{\strut \small{$checking\_status='<0'\,\,\,\land$}}\hbox{\strut \small{$duration\in [-0.45, 0.86]\,\,\, \land$}}\hbox{\strut \small{$installment\_commitment\geq-0.44$}}} & $0.83$ & $0.07$ & $0.88$ & $1.0$ \\
        \midrule
        \vtop{\hbox{\strut \small{$checking\_status='0<=X<200'\,\,\,\land$}}\hbox{\strut \small{$LSTAT\leq0.21 \land$}}\hbox{\strut \small{$duration\geq0.86 \land$}}} & $0.86$ & $0.06$ & $0.88$ & $1.0$ \\
    \end{tabular}
    }
    }
    \caption{Global results for the Boston dataset. Top: global discrepancy importances for each feature. Bottom: major discrepancy segments detected.}
    \label{fig:global_results_discussion}
\end{figure}

A natural follow-up  question to these local observations concerns their generalization to other instances. We investigate this issue by aggregating (average value) the ranges of all of the intervals generated by DIG, defining a notion of feature importance at the classifier pool level. The higher this importance is, the larger the discrepancy area is along this feature, meaning that the considered models are more uncertain about where to split classes along this feature.
The results, presented in Figure~\ref{fig:global_results_discussion} on the left for continuous features, confirm that all features are not equally responsible for discrepancies. In particular, concurring with our observations at the local level, the two considered models seem to have the most trouble deciding on where to separate classes along the feature describing the duration of the loan ($duration$) or the age of the customer, indicating to the model developer that these features may carry out some uncertainty and its effect on the predicted attributed may benefit from more investigation.

Another interesting global information consists in understanding \emph{where} in the feature space the discrepancy regions reside. We propose to answer this question by training a decision tree on the nodes generated by DIG, labelled with~$\mathbbm{1}_{\mathcal{D}_{\mathbb{F}_{\epsilon}}}(x)$, to identify discrepancy segments. Although describing these segments in regions that do not contain any ground-truth data may be desirable, in this context, we suppose that the Data Scientist would like to focus on areas of the future space supported by a minimum of training instances.
As traditional decision tree algorithms may lead to degenerate solutions in this context, e.g. containing DIG nodes but little to no actual ground-truth data, we use a simple custom algorithm\footnote{The algorithm simply adds a new stopping criterion to the CART algorithm: the number of \emph{ground-truth} instances covered by a leaf, i.e. instances that are not used to train the tree. Indeed the tree is trained on the nodes generated by DIG.} based on CART~\cite{breiman1984classification}. 
This enables the generation of global human-understandable discrepancy segments, at the cost of precision in the discrepancy prediction. In the considered example, two major discrepancy segments have been identified, described in the right table in Figure~\ref{fig:global_results_discussion}, accounting in total for $35\%$ of the discrepancy nodes. Besides the description of the segment in the form of conjunctive rules (left column) and its associated average discrepancy score calculated over the DIG nodes falling into this segment (Disc. score), three additional information are provided: 'Expo' (for train exposition): the proportion of the training data that fit into this segment, indicating how in-distribution this segment is; 'F1(SVM)' and 'F1(XGB)': the F1 scores obtained for both classifiers, calculated over the training instances belonging to the segment.

A first observation is that these two detected segments cover here an important part of the training set (resp. $7\%$ and $6\%$ of the training data): the observed discrepancy is thus not a part of the feature space that is poorly represented in the data, and should thus be investigated quickly, as new instances are very likely to fall into these areas. Furthermore, we notice that the two considered models, despite achieving similar predictive performance over the test set, do not perform equally well in these segments: in both regions, the SVM classifier is largely outperformed by the XGBoost classifier. Besides implying to trust the XGBoost classifier rather than the SVM when a new prediction falls into these local areas, these observations suggest leads on how to increase the predictive performance of these models.
Beyond understanding the limits of the classifiers, the proposed approach thus allows to easily diagnose a source of discrepancies and take early remedial actions, such as prioritizing the sampling of new training instances in the identified segments or taking special precautions when a prediction is done in these regions of the space.
To solve this issue, a solution could be to ask for more observations in this part of the feature space, akin to
\emph{Query-by-committee}~\cite{Seung1992activecommittee}
strategy in active learning where instances to label are prioritized based on the disagreement of trained classifiers. In the mean time, another idea would be to involve a human in the loop to make the final decision when a new prediction is to be done in these regions.

In conclusion, understanding model discrepancies when modelling a task opens up several remediation steps to develop better ML systems.

\section{Conclusion}

In this paper we introduced and formalized the notion of prediction discrepancies, when several classifiers trained on the same data reach similar performances during test time despite having learned significantly different classification patterns. After showing the prevalence of the issue, we proposed DIG, an approach to address this issue with an effective algorithm.
The model-agnostic tool generates grounded and actionable explanations for the discrepancies impacting a prediction to enable the practitioner to take the best educated decision when selecting a model by anticipating its potential undesired consequences.
This work opens multiple perspectives, from further studying what causes discrepancies to proposing concrete actions to help the practitioner once DIG has helped understanding them. More generally, pursuing the work on problems and solutions for ML pipelines, in opposition to solely focusing on interpreting one model, constitutes a promising direction.

\newpage

\bibliography{biblio}

\newpage

\appendix

\section*{Appendix}
\label{sec:appendix}

\section{Proportion of predictions with discrepancies on datasets from OpenML-CC18 benchmark}

\resizebox{\columnwidth}{!}{
\begin{tabular}{ll}
\label{tab:app-survey}
{OpenML-CC18 Datasets} & Proportion of predictions with discrepancies \\
\midrule
pc4                                    &                                        16.7\% \\
pc3                                    &                                         9.3\% \\
jm1                                    &                          13.1\% \\
kc2                                    &                                        29.5\% \\
kc1                                    &                          11.6\% \\
pc1                                    &                          4.1\% \\
balance-scale                          &                                         0.3\% \\
mfeat-factors                          &                                         1.6\% \\
mfeat-fourier                          &                                        19.1\% \\
bank-marketing                         &                                        15.5\% \\
banknote-authentication                &                                         2.0\% \\
blood-transfusion-service-center       &                          30.1\% \\
cnae-9                                 &                                        10.0\% \\
first-order-theorem-proving            &                                        27.6\% \\
har                                    &                          1.8\% \\
ilpd                                   &                                        47.9\% \\
madelon                                &                                        18.4\% \\
nomao                                  &                          4.1\% \\
ozone-level-8hr                        &                                         4.9\% \\
phoneme                                &                                        15.7\% \\
qsar-biodeg                            &                          13.6\% \\
wall-robot-navigation                  &                                         0.8\% \\
breast-w                               &                                        70.8\% \\
semeion                                &                                         6.0\% \\
electricity                            &                                        15.2\% \\
wdbc                                   &                                         4.0\% \\
adult                                  &                          20.2\% \\
mfeat-karhunen                         &                                         2.4\% \\
mfeat-morphological                    &                          27.4\% \\
satimage                               &                          11.7\% \\
eucalyptus                             &                          60.2\% \\
mfeat-zernike                          &                          19.9\% \\
cmc                                    &                                        64.8\% \\
dresses-sales                          &                                        40.2\% \\
numerai28.6                            &                                        31.7\% \\
optdigits                              &                                         1.2\% \\
credit-approval                        &                          25.9\% \\
kr-vs-kp                               &                                         0.1\% \\
isolet                                 &                          2.9\% \\
vowel                                  &                                         1.6\% \\
credit-g                               &                                        30.8\% \\
pendigits                              &                                         0.5\% \\
diabetes                               &                                        26.3\% \\
sick                                   &                                         2.7\% \\
texture                                &                                         0.3\% \\
connect-4                              &                                        21.8\% \\
dna                                    &                                         5.4\% \\
churn                                  &                                         5.1\% \\
Devnagari-Script                       &                                         0.0\% \\
CIFAR\_10                               &                                        40.6\% \\
MiceProtein                            &                                         0.4\% \\
car                                    &                                         0.0\% \\
Internet-Advertisements                &                                         4.7\% \\
mfeat-pixel                            &                                         2.0\% \\
steel-plates-fault                     &                                        32.5\% \\
wilt                                   &                                         0.6\% \\
segment                                &                                         5.7\% \\
climate-model-simulation-crashes       &                          0.7\% \\
Fashion-MNIST                          &                                        19.3\% \\
jungle\_chess\_2pcs\_raw\_endgame\_complete &                                         8.4\% \\
Bioresponse                            &                          16.9\% \\
spambase                               &                                         8.6\% \\
PhishingWebsites                       &                          2.9\% \\
GesturePhaseSegmentationProcessed      &                                        36.3\% \\
analcatdata\_authorship                 &                                         0.4\% \\
splice                                 &                          11.8\% \\
analcatdata\_dmft                       &                          28.0\% \\
tic-tac-toe                            &                          0.7\% \\
vehicle                                &                                        16.0\% \\
mnist\_784                              &                                         9.4\% \\
letter                                 &                                         3.2\% \\
cylinder-bands                         &                                        38.3\% \\
\bottomrule
\end{tabular}
}

\section{Wasserstein distance to explore the structure of instances with prediction discrepancies}
\label{appendix:wasserstein}

\begin{figure}[h]
    \centering
    \includegraphics[width=\linewidth]{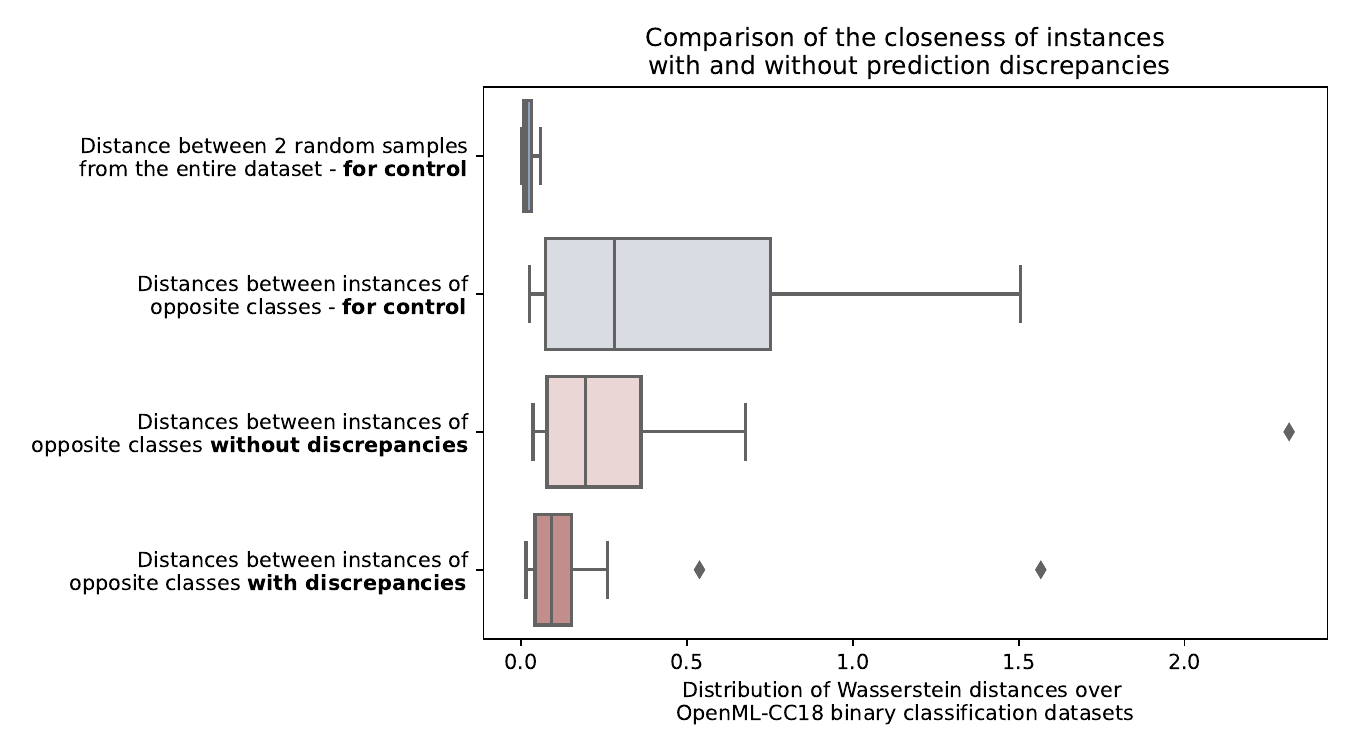}
    \caption{Wasserstein distance over subsets from opposite class of the instances with or without prediction discrepancies, over datasets of the OpenML-CC18 benchmarking suite. Instances with prediction discrepancies of opposite class are close in the feature space.}
\label{fig:wasserstein}
\end{figure}

We conduct an analysis on datasets of OpenML-CC18 to observe if there is a structure among instances with prediction discrepancies.
Our objective is now to observe if instances with prediction discrepancies are rather similar or dissimilar in the feature space $\mathcal{X}$.

To do so, for every dataset of OpenML-CC18 with a binary classification task, we compare (1) the distributions for each class of instances \textit{without} discrepancies (2) the distributions for each class of instances \textit{with} discrepancies.
The comparisons of the different pairs of distributions is done with the Wasserstein distance (or earth mover's distance, using \citet{pele2009}).
As "sanity checks" and baselines, we compute the Wasserstein distance (1) between 2 random samples drawn across the entire dataset is calculated (i.e. we expect distances close to 0 across datasets as the distributions resulting from this sampling should be very similar) and (2) between the distributions of instances of opposite classes no matter there discrepancy status across datasets (i.e. we expect a large distance as the distribution of instances of opposite classes in a classification problem should be well separated).

On Figure~\ref{fig:wasserstein}, we observe over the OpenML-CC18 datasets that the Wasserstein distances between instances of opposite classes \textit{with} prediction discrepancies are significantly lower than the distances between instances of opposite classes \textit{without} prediction discrepancies.
We can derive two hypotheses from that observation: (1) instances with prediction discrepancies are rather similar and may be grouped in the feature space (2) a significant source of prediction discrepancies may be instances of opposite classes that are too close for the models to be able to properly separate them.



\section{Experimental Protocol}
\label{sec:appendix-experimental-protocol}

This section gives more insight on how the experiments were conducted, for the sake of transparency and reproducibility.

\subsection{Dataset Descriptions}

Table~\ref{tab:dataset-descr}: Description of the tabular datasets considered, as well as the performance obtained with the Autogluon pool (for the experiments of Section~\ref{sec:evaluation}). The pools are further described in the next section. 

6 tabular datasets are considered, chosen for their numerical attributes and often used in the interpretability literature: the toy dataset half-moons
\footnote{\url{https://scikit-learn.org/stable/modules/generated/sklearn.datasets.make\_moons}}
, the Breast Cancer dataset~\cite{BreastCancerDataset}, the Boston dataset~\cite{BostonDataset} (transformed into a binary classification problem), 
the Churn dataset\footnote{\url{https://epistasislab.github.io/pmlb/profile/churn.html}}, the Online News Popularity dataset~\cite{NewsDataset}, and the Adult dataset\footnote{\url{https://archive.ics.uci.edu/ml/datasets/adult}}.
After dropping categorical features (for experiment 1) and  normalizing the data, each dataset is split between training and testing sets (2/3-1/3 split).

Notations
\begin{itemize}
    \item $n$: number of training instances kept
    \item $d$: number of original features
\end{itemize}

\begin{table}[ht]
    \centering
    \begin{tabular}{l |c|c|c}
        Dataset & $n$ & $d$  & $\mathbb{F}_{\epsilon}$ accuracies\\
        \toprule
        half-moons & $1000$ & $2$ & $0.90-0.92$ \\
        
        boston & $506$ & $19$ & $0.90-0.92$ \\
        
        breast-cancer & $569$ & $30$ & $0.94-0.95$ \\
        
        churn & $5000$ & $21$ & $0.80-0.82$ \\
        
        news & 39797 & $61$ & $0.81-0.83$\\
        
        adult & $48842$ & $14$ & $0.61-0.66$\\
        
        german & $1000$ & $20$ & $0.80-0.85$\\
        \bottomrule
        
    \end{tabular}
    \caption{Description of the datasets. In the right columns are indicated the range of the predictive performances (F1 score) obtained with the AutoGluon pool used for each experiment.}
    \label{tab:dataset-descr}
\end{table}

\subsection{Pool used}

The implementation of the $\epsilon$-comparable pool of classifiers is especially important to evaluate the efficiency of the approaches, as it directly influences the discrepancy areas that are to be explained.
Although the proposed method can be used for any pool, one motivation for this work refers in particular to models with similar predictive performance ($\epsilon$-comparable pools of classifiers).
In order to make this choice as transparent as possible, we use a standard automated machine learning library, Autogluon~\cite{autogluon}.
This auto-ML tool allows to optimize model performances by training multiple models with different parameters and pre/post processing.
We build $\epsilon$-comparable pools of classifiers by selecting each time the most performing models, with~$\epsilon=0.05$ for the $\epsilon$-comparable pools of classifiers.
The values chosen for the parameters of the pools are the following:

\begin{itemize}
    \item \textbf{AutoGluon:} Default parameters.
    \item \textbf{Auto-sklearn:} Default parameters.
    \item \textbf{Basic pool}: An ensemble of random forest (200 trees, maximum depth 10), SVM (RBF kernel), Logistic Regression (no penalization) and, XGBoost (200 estimators) classifiers, implemented with the scikit-learn library. Parameters not mentioned are left untouched.
\end{itemize}

For all pools and all experiments, we set: $\epsilon=0.05$ and leave out the classifiers that are not $\epsilon$-equiperforming (calculation over the validation set).

\subsection{General protocol.}
Once the pool classifiers are fitted on the training set, we use the first step of DIG to learn a graph~$\mathcal{G}$ to discover discrepancies between models of the pool.
The values chosen for parameters~$k$ and $n\_{epochs}$ are indicated in each experiment and in Appendix.
For each dataset, the same pool was used across all experiments.

\subsection{DIG parameters.}

In this section are described the values considered for the hyperparameters of the proposed algorithm, DIG. The two parameters are $k$, the number of neighbors used to initially instantiate the graph, and $n_{epochs}$, the number of learning iterations. Values are shown in Table~\ref{tab:dig-params1} and~\ref{tab:dig-params2}. 

\subsubsection{Experiment 1  (Sec.~\ref{sec:experiment-elephant})}
\label{sec:appendix-expe1}

\begin{table}[h!]
    \centering
    \begin{tabular}{l|c|c}
        Dataset & $k$ & $n_{epochs}$ \\
        \midrule
        half-moons & 10 & 5\\
        boston & 50 & 5 \\
        breast-cancer & 50 & 5\\
        churn & 500 & 10 \\
        news & 500 & 10 \\
        adult & 100 & 10 \\
        german & 30 & 5 \\
        \bottomrule
    \end{tabular}
    
    \caption{Value set for the parameters of Algorithm~\ref{alg:pool2graph-preprocessing} for Experiment 1.}
    \label{tab:dig-params1}
\end{table}

Parameter values for DIG chosen for experiment 1. The parameter values for the competitors are described in Section~\ref{sec:exeriment1-blackbox}.

\subsubsection{Experiment 2 (Sec.~\ref{sec:experiment-precision})}
\label{sec:appendix-expe2}

\begin{table}[ht]
    \centering
    \begin{tabular}{l|c|c}
        Dataset & $k$ & $n_{epochs}$ \\
        \midrule
        half-moons & 10 & 40 \\
        boston & 10 & 15 \\
        breast-cancer & 10 & 15 \\
        churn & 10 & 15 \\
        news & 10 & 15 \\
        \bottomrule
    \end{tabular}
    \caption{Value set for the parameters of Algorithm~\ref{alg:pool2graph-preprocessing} for Experiment 2.}
    \label{tab:dig-params2}
\end{table}
Additionally, as mentioned, for each interval, we set the number of instances generated to estimate using Monte-Carlo the precision of the interval to $100$.

\subsubsection{Experiment 3 (Sec.~\ref{sec:usecases})}
\label{appendix:usecases}

Similarly to previous experiments, a pool of classifiers is trained using AutoGluon (default parameters). The F1 score on test data reached was between $0.80$ and $0.85$. In the test set, $32.5\%$ of the instances are assigned with conflicting predictions. Values for the parameters of Alg~2 are shown in Table~\ref{tab:dig-params3}.
\begin{table}[ht]
    \centering
    \begin{tabular}{l|c|c}
        Dataset & $k$ & $n_{epochs}$ \\
        \midrule
        German Credit & 30 & 5\\
        \bottomrule
    \end{tabular}
    \caption{Value set for the parameters of Algorithm~\ref{alg:pool2graph-preprocessing}.}
    \label{tab:dig-params3}
\end{table}

The number of intervals to extract in Algorithm~\ref{alg:pool2graph-inference} was set to 1.

\section{Additional Results}

\subsection{Additional results for Experiment 1}
\label{appensix:results-expe1}

2D visualization of the sampling strategies for mode competitors: a Gaussian Mixture Model and a Variational Autoencoder. The number of components of the GMM was chosen by cross-validation. The VAE is composed of an encoder, built using fully connected layers of respective dimensions 64, 32, 5 (dimension of the latent space); and a decoder of dimensions 34 and 64.

\begin{figure}[h]
    \centering
    \includegraphics[width=0.44\linewidth]{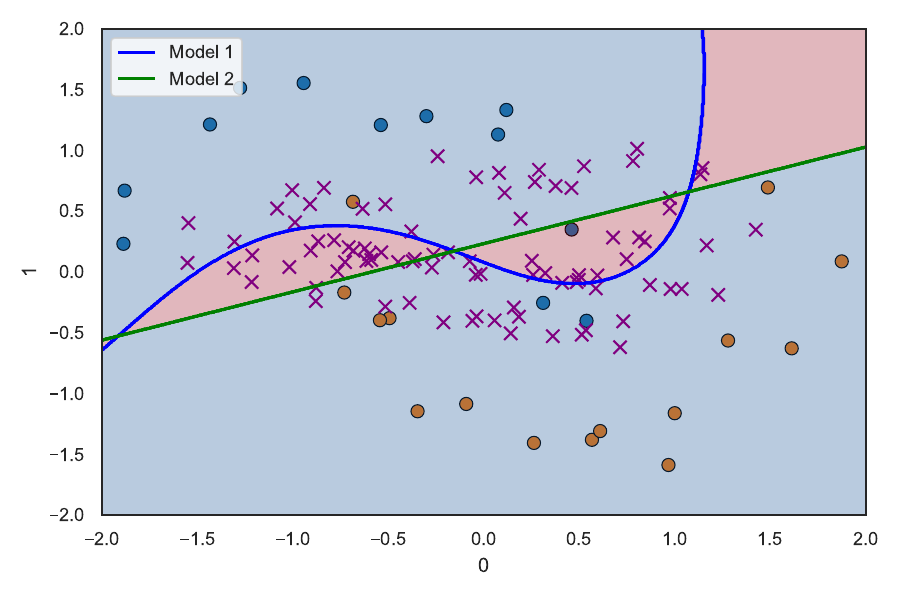}
    \includegraphics[width=0.44\linewidth]{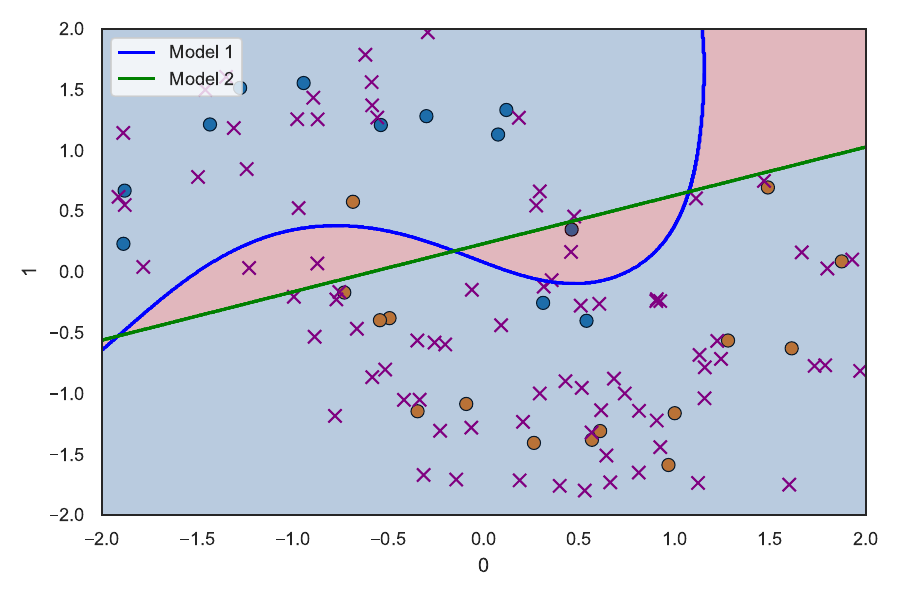}
    \includegraphics[width=0.44\linewidth]{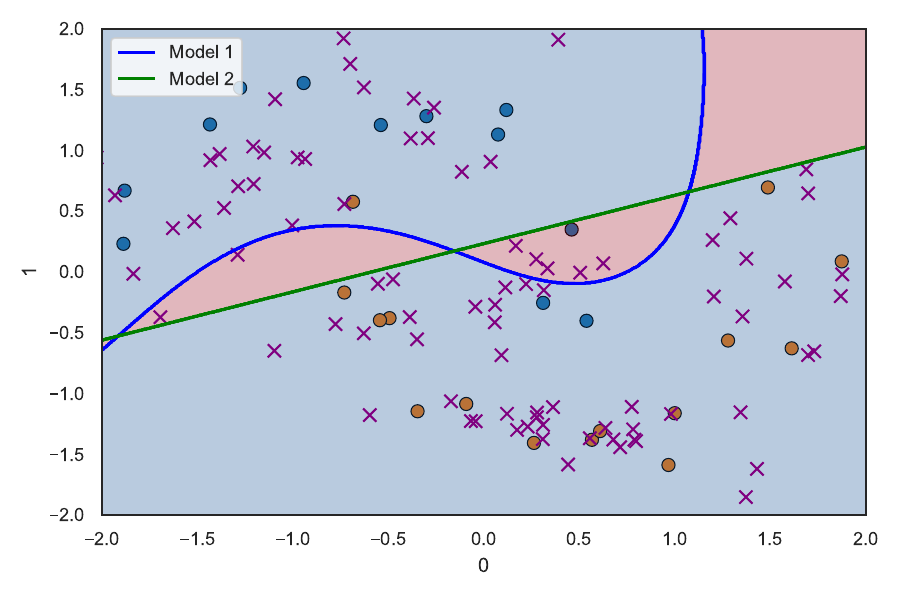}
    \includegraphics[width=0.44\linewidth]{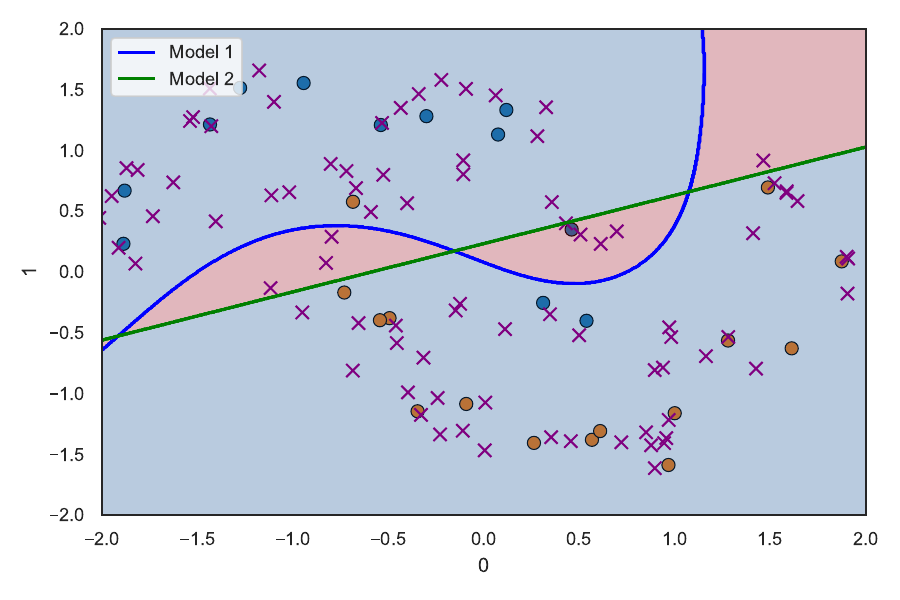}
    \caption{Sampling strategies of (from left to right, top to bottom): DIG, KDE, GMM and VAE}
    \label{fig:results-toyscenario-halfmoons}
\end{figure}

Additional results for Experiment 1 (Section~\ref{sec:experiment-elephant})

\begin{tabular}{l|c|c|c|c}
    Dataset & DIG & KDE & GMM & VAE \\
    \midrule
    half-moons & $\textbf{0.96\,(0.02)}$ & $0.92\,(0.03)$ & $0.94\,(0.02)$ & $0.92\,(0.02)$\\ 
    
    boston & $\textbf{0.78\,(0.05)}$ & $0.57\,(0.07)$ & $0.68\,(0.08)$& $0.45\,(0.03)$\\
    
    breast-cancer & $\textbf{0.75\,(0.05)}$ & $0.40\,(0.02)$ & $0.50\,(0.09)$& $0.55\,(0.08)$\\
    
    churn & $\textbf{0.60\,(0.02)}$ & $0.59\,(0.01)$ & $0.61\,(0.04)$& $0.60\,(0.01)$\\

    news & $\textbf{0.60\,(0.02)}$ & $0.42\,(0.05)$  & $0.50\,(0.05)$& $0.51\,(0.07)$\\
    
    adult & $\textbf{0.81\,(0.03)}$ & $0.60\,(0.02)$  & $0.62\,(0.03)$ & $0.62\,(0.04)$\\
    
    german & $\textbf{0.71\,(0.03)}$ & $0.65\,(0.02)$  & $0.60\,(0.03)$& $0.63\,(0.03)$ \\
    
    \bottomrule
\end{tabular}

\subsection{Experiment: handling categorical attributes}
\label{appensix:results-categorical}

Results for Experiment 1 using Gower's distance~\cite{gower1971general}. The implementation used is from the python package \emph{gower}\footnote{https://pypi.org/project/gower/}.

Results for Adult dataset: $0.79 (0.04)$
Results for German credit dataset: $0.72 (0.01)$

\subsection{Additional Explanation Examples}
\label{appendix:results-output}



\begin{figure}[h]
    \centering
    \includegraphics[width=0.6\linewidth]{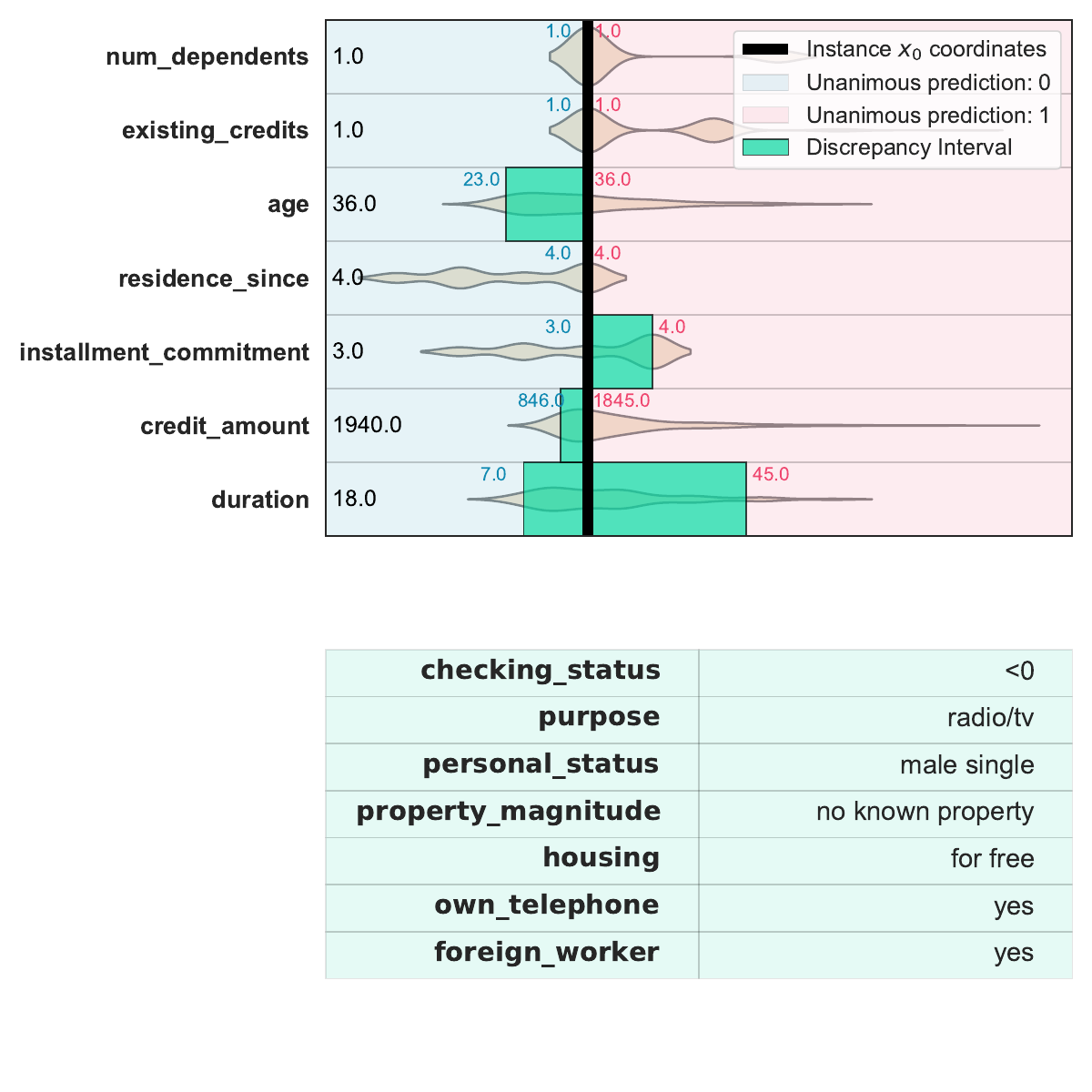}
        
    \caption{Output of DIG for an instance of the German credit dataset.}
    \label{fig:dig-output-a1}
\end{figure}

\begin{figure}[h]
    \centering
    \includegraphics[width=0.6\linewidth]{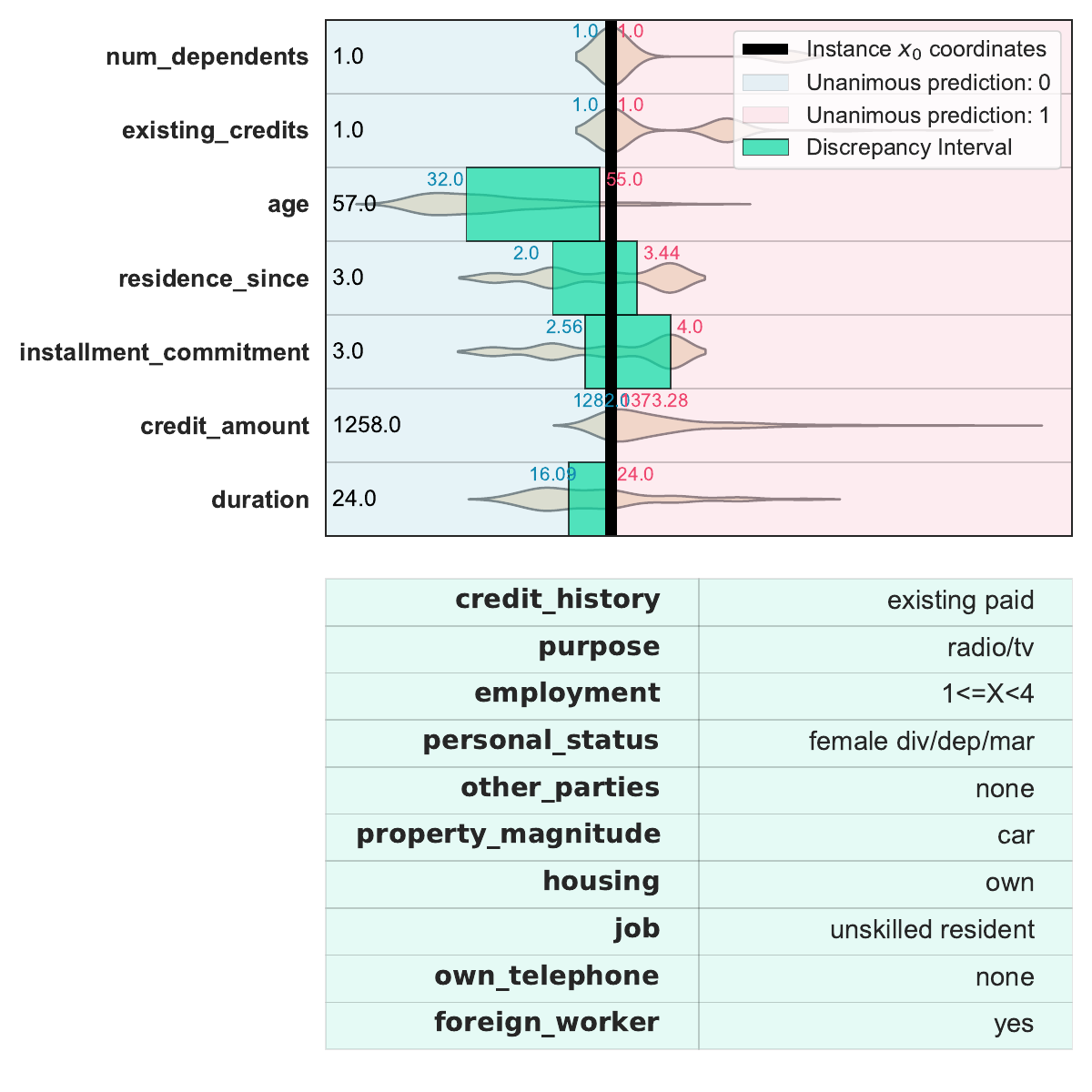}
        
    \caption{Output of DIG for an instance of the German credit dataset.}
    \label{fig:dig-output-a2}
\end{figure}

Figures~\ref{fig:dig-output-a1} and~\ref{fig:dig-output-a2} show the output obtained with DIG for additional instances from the German Credit Dataset.


\subsection{Examples of outputs for image classification}
\label{appendix:digcv}

 In this section, we show the first results that were obtained by our method on two datasets, MNIST and FashionMNIST, on which we train two classifiers, a convolutional network 
 and a SVM classifier.
Although the accuracy difference between these models is relatively small ($0.01\%$ on MNIST and $0.03$ on FMNIST), we measure that the models are disagreeing respectively over $2.23\%$ and $XX$ of the instances of the tests sets.

\begin{figure}[h]
    \centering
    \includegraphics[width=0.7\linewidth]{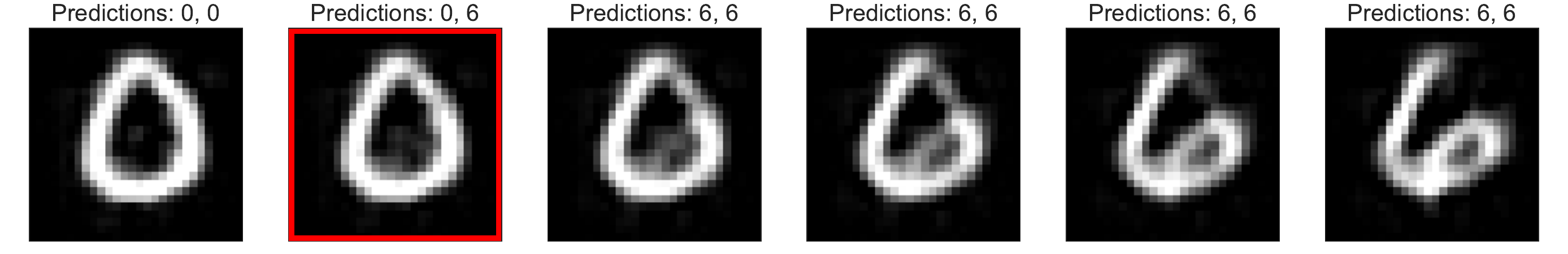}
    \includegraphics[width=0.7\linewidth]{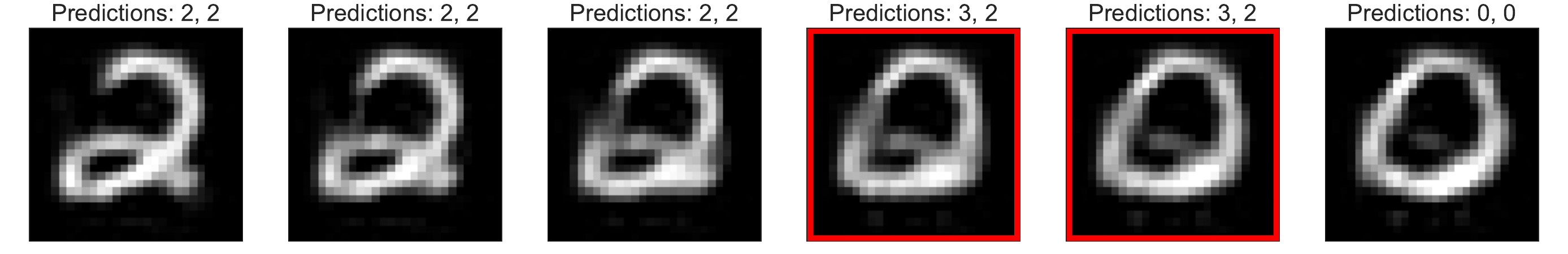}
    \includegraphics[width=0.7\linewidth]{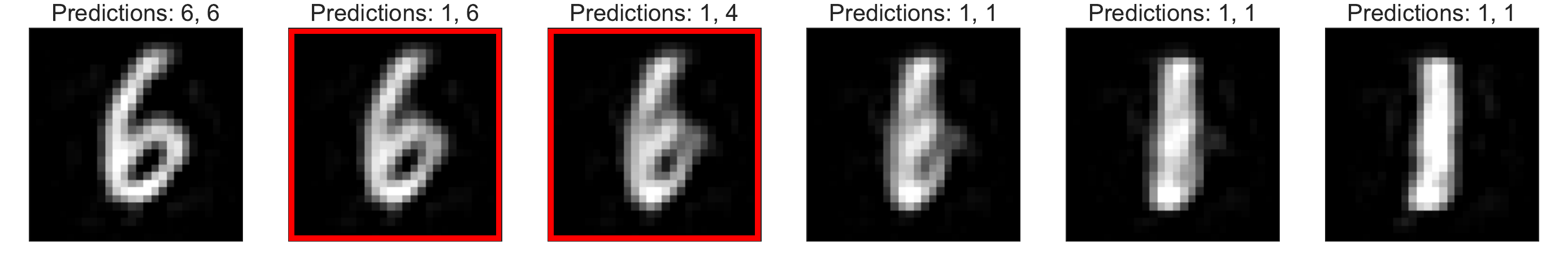}
    \caption{Results obtained for three instances of MNIST.}
    \label{fig:results-MNIST}
\end{figure}

\begin{figure}[h]
    \centering
    \includegraphics[width=0.8\linewidth]{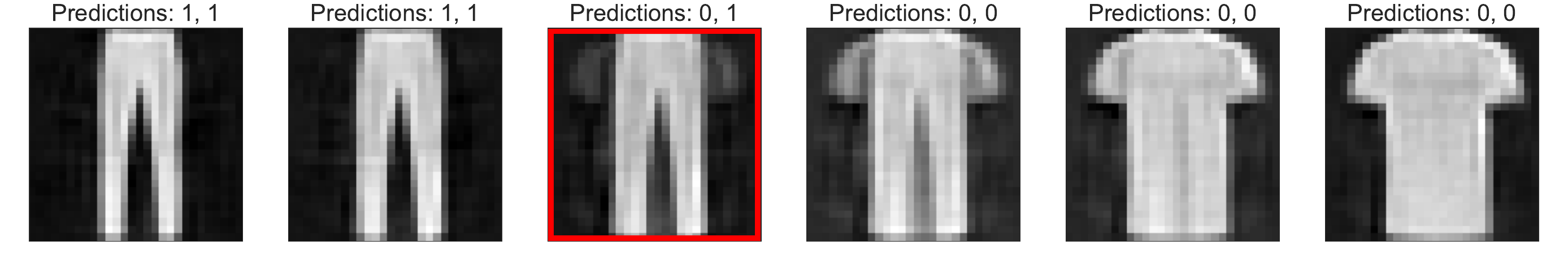}
    \includegraphics[width=0.8\linewidth]{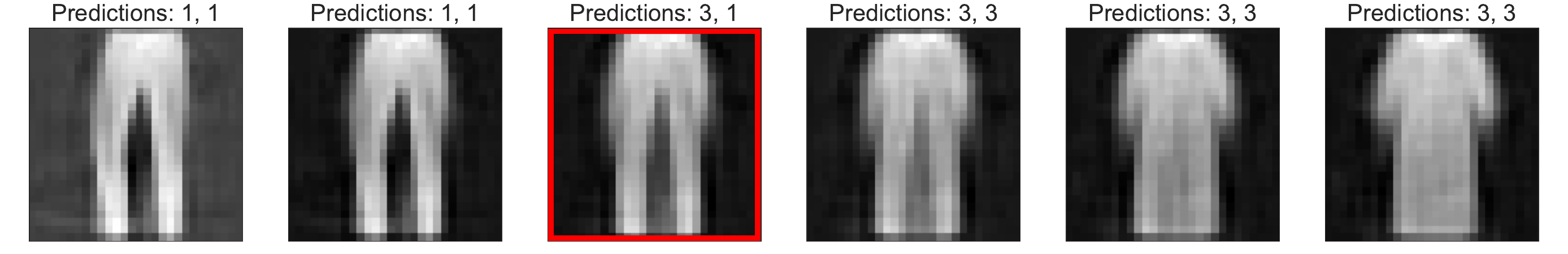}
    \includegraphics[width=0.8\linewidth]{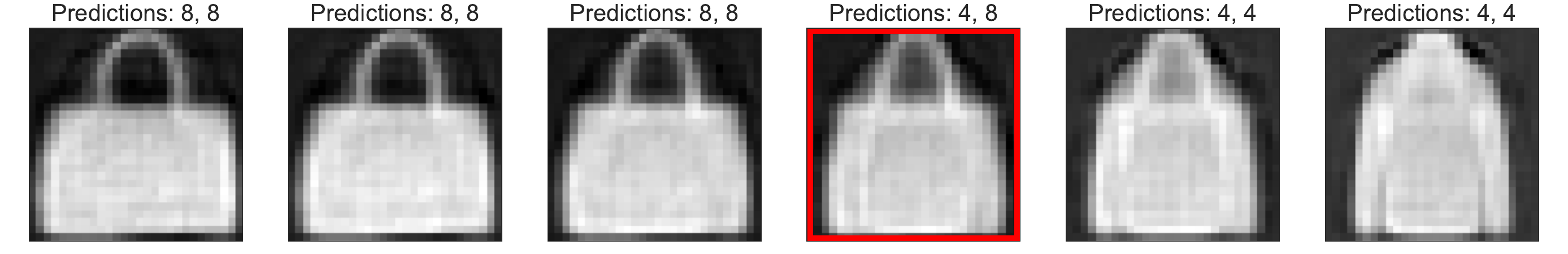}
    \caption{Results obtained for three instances of FashionMNIST.}
    \label{fig:results-FMNIST}
\end{figure}
 
To investigate these disagreements, we apply DIG-CV to both datasets
and show some illustrative results in Figures~\ref{fig:results-MNIST} and~\ref{fig:results-FMNIST}. On the extreme left and right columns of each row are shown the reconstruction $x'$ of two instances $x$ from the training set. In-between are shown reconstructions $g_{\theta'}(I)$ of the instances sampled in the latent space~$Z$ by DIG. The instances whose reconstruction falls in a discrepancy area are highlighted using a red square around the image.
Using these explanations, the practitioner can detect uncertain areas of the feature space, and depending on the case, perform remediating actions. Such actions include for instance labelling of more data in these uncertain areas, or asking for a human in the loop to perform the decision if models can not agree.

\end{document}